\pdfoutput=1

\PassOptionsToPackage{dvipsnames,svgnames,table}{xcolor}
\documentclass[11pt]{article}
\usepackage{authblk}
\usepackage[]{EMNLP2023}

\usepackage{todonotes}
\usepackage{adjustbox}
\usepackage{booktabs}
\usepackage{graphicx}
\usepackage{enumitem}
\usepackage{xcolor}
\usepackage{multicol}
\usepackage{amsmath}
\usepackage{bold-extra}
\usepackage{subcaption}
\usepackage{float}
\usepackage{nicefrac}
\usepackage{color, colortbl}
\usepackage{times}
\usepackage{latexsym}
\usepackage{makecell}
\usepackage{amssymb}
\usepackage{multirow}
\usepackage{tablefootnote}
\usepackage{textcomp}
\usepackage{pifont}
\usepackage[dvipsnames]{xcolor}
\usepackage{array}
\usepackage{booktabs}
\usepackage{authblk}
\usepackage{CJKutf8}
\usepackage{array}
\usepackage{longtable,supertabular}
\usepackage{kotex}
\usepackage{amssymb}
\usepackage{pifont}

\usepackage{url}
\renewcommand{\footnotesize}{\fontsize{8.5pt}{10.5pt}\selectfont}

\newcommand{\mytilde}{\raise.17ex\hbox{$\scriptstyle\mathtt{\sim}$}}

\usepackage[T1]{fontenc}

\usepackage[utf8]{inputenc}

\usepackage{microtype}

\usepackage{inconsolata}

\title{Low-Rank Adaptation  for Multilingual Summarization: \\
An Empirical Study}

\author{
    \textbf{Chenxi Whitehouse$^{1,2,}$\thanks{$^*$Work conducted as Research Intern at Google DeepMind.}\quad Fantine Huot$^{2}$\quad Jasmijn Bastings$^2$} \\ \textbf{Mostafa Dehghani$^{2}$\quad Chu-Cheng Lin$^{3}$\quad Mirella Lapata$^{2}$ }\\
    $^{1}$University of Cambridge \: $^{2}$Google DeepMind \: $^{3}$Google   \\
    \texttt{chenxi.whitehouse@cl.cam.ac.uk} \\
  \texttt{\{fantinehuot, bastings, dehghani, kitsing, lapata\}@google.com}
}

\begin{document}
\maketitle
\begin{abstract}

Although the advancements of pre-trained Large Language Models have significantly accelerated recent progress in NLP, their ever-increasing size poses significant challenges for conventional fine-tuning, especially in memory-intensive tasks. We investigate the potential of Parameter-Efficient Fine-Tuning, focusing on Low-Rank Adaptation (LoRA), in the domain of multilingual summarization, a task that is both challenging (due to typically long inputs), and relatively unexplored.  We conduct an extensive study across different data availability scenarios, including high- and low-data settings, and cross-lingual transfer, leveraging models of different sizes. Our findings reveal that LoRA is competitive with full fine-tuning when trained with high quantities of data, and excels in low-data scenarios and cross-lingual transfer. We also study different strategies for few-shot cross-lingual transfer, finding that continued LoRA tuning outperforms full fine-tuning and the dynamic composition of language-specific LoRA modules.
\end{abstract}

\section{Introduction}
\label{sec:introduction}

The emergence of pre-trained Large Language Models (LLMs), such as
PaLM~2 \cite{anil2023palm}, LLaMA~2 \cite{touvron2023llama2}, and the
GPT family from OpenAI, has significantly advanced the state of the
art in numerous NLP applications. However, the expansion in the size of LLMs poses significant challenges for traditional fine-tuning, particularly when faced with many downstream tasks or tasks with a large memory footprint, e.g.,~due to processing long inputs.

Parameter-Efficient Fine-Tuning (PEFT) methods have recently shown
promise to adapt a pre-trained model to different tasks by
selectively fine-tuning a small subset of additional parameters.
Widely-adopted PEFT techniques include adapters
\cite{pmlr-v97-houlsby19a, pfeiffer-etal-2021-adapterfusion}, Low-Rank
Adaptation (LoRA; \citealt{hu2022lora}), \mbox{prefix-tuning}~\cite{li-liang-2021-prefix}, and \mbox{prompt-tuning}~\cite{lester-etal-2021-power}. Among these, LoRA has
become one of the most popular approaches, achieving state-of-the-art performance
without introducing latency at inference time.  The majority of PEFT
studies have focused on natural language understanding,
e.g.,~classification tasks as exemplified in the GLUE
\cite{wang-etal-2018-glue} and SuperGLUE \cite{wang2019superglue}
benchmarks, and monolingual generation, e.g., table-to-text generation
or summarization \cite{li-liang-2021-prefix}.

In this paper, we empirically investigate the potential of LoRA in the
domain of \emph{multilingual} summarization, a task that is both challenging
and relatively unexplored.  Multilingual summarization often involves
processing lengthy inputs \cite{hasan-etal-2021-xl}, providing a
natural testbed for the effective use of PEFT methods.  In addition to
being able to understand long documents, models are expected to fluently 
generate sentences in many languages, requiring significant
linguistic versatility.  Multilingual tasks face additional challenges
pertaining to the availability of resources (e.g.,~for training). It
is unrealistic to expect that large-scale and high-quality data will
be available or created for every language
\cite{parida-motlicek-2019-abstract}.  In scenarios where
multilingual data is scarce, PEFT methods which selectively update a
small number of parameters seem more suitable while fine-tuning can
lead to overfitting or catastrophic forgetting \cite{PMID:28292907, mitchell2022fast}.

This motivates us to explore the following research questions: (i)~Can
LoRA be effectively applied to complex multilingual summarization
tasks?  and (ii)~Under which conditions does LoRA exhibit the most
potential?  To answer these questions, we investigate different data
availability scenarios: \mbox{\emph{high-data}} regime (high
quantities of training data are available for all languages),
\mbox{\emph{low-data}} regime (training data is limited but available
for all languages), and \emph{cross-lingual transfer} (zero or only a
few examples are available for some languages).  In the latter case, a model trained on a high-resource language (e.g.,~English) is localized
to additional languages for which data is scarce or unavailable
\cite{artetxe-etal-2020-cross, K2020Cross-Lingual}. In addition to
mimicking real-world conditions, the cross-lingual transfer setting
allows us to experiment with the composition of language-specific LoRA
modules, including the recently proposed few-shot LoraHub
\cite{huang2023LoraHub}.  Our experiments are conducted on two multilingual
summarization datasets, XLSum \cite{hasan-etal-2021-xl} and XWikis
\cite{perez-beltrachini-lapata-2021-models}, using different sizes of
the PaLM~2 model, an LLM trained on multilingual text spanning more
than 100~languages \cite{anil2023palm}.

To summarize, our contributions are as follows: (i)~we conduct a comprehensive study of the effectiveness of LoRA for multilingual
summarization under different data regimes; (ii) we showcase the
benefits of LoRA in low-data and cross-lingual transfer settings; and
(iii)~we investigate how to best leverage LoRA for cross-lingual
transfer subject to the availability of target language examples.

\section{Related Work}
\label{sec:related-work}

\paragraph{Parameter Efficient Fine-Tuning \textnormal{methods}} aim to enhance
computational efficiency while maintaining competitive performance
compared to full fine-tuning.  LoRA is one of the most popular PEFT
approaches \cite{hu2022lora, chen-etal-2022-revisiting}. It reduces
the number of trainable parameters by learning pairs of
rank-decomposition matrices while freezing the model's original
weights. This vastly reduces storage requirements for large language
models adapted to specific tasks and enables efficient task-switching
during deployment, without introducing inference latency. More recent
work explores how to adaptively adjust the rank of the matrices
\cite{zhang2023adaptive, valipour-etal-2023-dylora}, proposes
generalizations of LoRA and related PEFT approaches under a common
framework \cite{he2022towards, chavan2023one}, and combines LoRA with
quantization \cite{dettmers2023qlora}.  However, most of these studies
focus on classification and monolingual generation tasks. In contrast,
we investigate the potential of LoRA in the domain of multilingual
summarization, a task that is both challenging and relatively
unexplored.

\paragraph{Cross-lingual Transfer \textnormal{requires}} a model to learn a task
from labeled data in one language (typically English), and then
perform the equivalent task in another language where no or very
little labeled data is available \cite{artetxe-etal-2020-cross, K2020Cross-Lingual, lauscher-etal-2020-zero, whitehouse-etal-2022-entitycs, whitehouse-etal-2023-llm}. Previous studies focusing on PEFT methods for cross-lingual transfer have explored
adapter-based approaches \cite{pfeiffer-etal-2020-mad, ansell-etal-2021-mad-g} and composable sparse fine-tuning \cite{ansell-etal-2022-composable}, among others.
\citet{vu-etal-2022-overcoming} evaluate prompt-tuning
\cite{lester-etal-2021-power} in a zero-shot setting for cross-lingual
summarization, focusing on the Wikilingua dataset
\cite{ladhak-etal-2020-wikilingua}. Their study does not cover LoRA,
nor does it explore scenarios with more available data (e.g.,~few-shot
settings).

\label{sec:weight_composition}
\paragraph{Model Composition and Weight Merging \textnormal{aim}} to  enable generalization
to unseen tasks by combining individually trained models.  Previous
work includes weight composition guided by task similarity
\cite{lv-etal-2023-parameter} or arithmetic operations such as
addition or subtraction \citep{zhang2023composing}, multi-task prompt
pre-training \cite{sun-etal-2023-multitask}, and combining models in
parameter space by minimizing prediction differences between a merged
model and individual models \citep{jin2023dataless}. For our
multilingual summarization task, we also explore the composition of
language-specific LoRA matrices through weight averaging, as well as
dynamic weight composition when few-shot samples are available
\cite{huang2023LoraHub}.

\section{LoRA for Multilingual Summarization}
\label{sec:lora-mult-summ}

We now present the fundamentals of LoRA \cite{hu2022lora}
and then discuss how individual LoRA modules can be combined
\cite{huang2023LoraHub} for cross-lingual transfer. We also introduce
our assumptions regarding the availability of training data in the domain of multilingual summarization.

\subsection{LoRA and LoraHub}
\label{sec:lora}

\paragraph{LoRA} Let~\mbox{\(W_{0}\in\mathbb{R}^{d \times k}\)}
denote the weight matrix of a pre-trained LLM (where~$d$ is the input
dimension and $k$~is the output dimension). The key idea of LoRA is to
represent the fine-tuned~$W$ with a low-rank decomposition $W_0+ \Delta W= W_0 +BA$, where
\(B\in\mathbb{R}^{d \times r}\) and~\(A\in\mathbb{R}^{r \times k}\),
and $r \ll \min(d,k)$, making $BA$ a low-rank matrix compared to
$W_0$. During training, $W_0$ is frozen, while $B$ and $A$ contain
trainable parameters which are effectively a portion (\(2r/d\)) of the
parameters compared to full fine-tuning. Although LoRA can be in
principle applied to any subset of weight matrices, \citet{hu2022lora}
only update the weight matrices in the \mbox{self-attention} module of the
Transformer architecture. We also follow this recipe in experiments
and update all four attention matrices (i.e.,~\textit{query},
\textit{key}, \textit{value}, and \textit{out}).

\paragraph{LoraHub} is a gradient-free, few-shot learning approach,
recently proposed by \citet{huang2023LoraHub}. It focuses on composing
individually trained LoRA modules for cross-task
generalization. Available LoRA modules $m_i$ are synthesized into module
\mbox{$\hat{m} = \sum_{i=1}^{N} w_i m_i$} where $w_i$ is
a scalar weight that can assume positive and negative values. The
optimal weighted sum is learned through black-box gradient-free
optimization \cite{pmlr-v162-sun22e}, based on performance metrics on
a few examples representative of a new target task.

\subsection{Data Regimes}
We investigate the effectiveness of LoRA for multilingual
summarization under the following data assumptions:

  \paragraph{High Data} This scenario assumes that 
  \emph{sufficient} training data is available in all languages of
  interest. Such data could be obtained through automatic pipelines or
  crowdsourcing.

  \paragraph{Low Data} In this scenario, we assume 
  that a \emph{limited} number of examples are available in the target
  languages of interest, typically in the order of dozens or a few
  hundred. This scenario is common when working with low-resource
  languages or when data cannot be easily obtained through
  crowdsourcing but requires input from expert annotators.
  
   \paragraph{Cross-Lingual Transfer} Within this context, we consider
   scenarios where training examples are primarily available in one or
   a few high-resource languages. We explore three settings
   corresponding to the following assumptions: (i) only English training data is available; (ii)
   training data is available in some languages besides English, which
   creates a more complex multilingual setting; and~(iii)~a small number
   of labeled examples are available in the target language, allowing
   us to study few-shot cross-lingual generalization.

\begin{table}[!t]
\centering

\addtolength{\tabcolsep}{-2.5pt}
\scalebox{0.75}{
\begin{tabular}{l|cc}
\toprule
\textbf{Dataset}& \sc \textbf{XLSum} &  \sc \textbf{XWikis} \\
\midrule
 Source &  BBC News  &  Wikipedia \\
 Languages & 44 & 5 \\
 Train/Val/Test Data & 1.12M / 114K / 114K  &  1.43M / 40K / 35K\\
Input/Output Words &  470.2 / 22.1  & 1042.7 / 63.7 \\

\bottomrule
\end{tabular}}
\caption{Summary statistics for the XLSum and XWikis multilingual
  summarization datasets.  Train/Val/Test shows the number of examples
  in each split.  Input/Output shows the average number of \textit{words}
  in the \textit{English} input document and output summary. XWikis
  has long documents and multi-sentence summaries.}
\label{tab:stats}
\end{table}

\section{Experimental Setup}
\label{sec:experimental-setup}
This section introduces the datasets and models used in our study. We
further elaborate on the details of our experimental setup, and the
metrics used to assess the generated summaries.

\subsection{Datasets}
We perform experiments on two multilingual abstractive summarization
datasets which differ with respect to the number of languages they cover,
the number of data samples available, and the summarization task
itself (short vs long summaries).  Dataset statistics are presented
in \autoref{tab:stats}.

\paragraph{XLSum}\hspace*{-.25cm}\cite{hasan-etal-2021-xl} contains
over one million article-summary pairs in 45 languages.  The dataset
was automatically collated from BBC News, under the assumption that
the introductory sentence in the article is effectively a summary of its content.  The number of training examples varies significantly
among languages, with English having more than~300K instances, and
Scottish-Gaelic just above~1K (see \autoref{tab:xlsum-data} in
\autoref{sec:xlsum}).

\paragraph{XWikis}\hspace*{-.25cm}\cite{perez-beltrachini-lapata-2021-models}
consists of document-summary pairs with long documents and
multi-sentence summaries. It was synthesized from Wikipedia articles,
under the assumption that the body of the article and its lead
paragraph together form a document-summary pair.  XWikis covers five
languages, i.e.,~Czech, German, English, French, and Chinese.  It also
includes cross-lingual document-summary instances, created by
combining lead paragraphs and article bodies from Wikipedia titles
that are language-aligned.  In our experiments, we focus on cases
where the article and the summary are in the \emph{same} language (see
\autoref{tab:xwikis-data} in \autoref{sec:xlsum}).

\subsection{Modeling Details}
\label{sec:modelling-details}

Our experiments focus on PaLM~2 \cite{anil2023palm}, a decoder-only
LLM which, compared to PaLM \cite{JMLR:v24:22-1144}, exhibits superior
multilingual and reasoning capabilities, as well as better
compute efficiency.  Specifically, we employ two sizes (XXS and S) of
the instruction-tuned FLAN-PaLM 2 model \cite{wei2022finetuned}.  All
experiments were conducted on cloud TPUs,\footnote{See \autoref{sec:tpu} for more details of the TPUs used.} with a learning rate in the
range of \{$1e^{-3}$, $2e^{-4}$, $2e^{-5}$\}.  The input/output length
was truncated at~2,048/128 tokens for XLSum and~2,048/256 for XWikis.

\subsection{Automatic Evaluation}
\label{sec:evaluation}

We evaluate the quality of the generated summaries along three
dimensions, namely \textit{relevance}, \textit{faithfulness}, and \textit{conciseness}. In terms
of relevance, we employ the widely used ROUGE score
\cite{lin-2004-rouge}, which measures the degree of n-gram overlap
between generated summaries and reference text.  Following
\citet{aharoni-etal-2023-multilingual}, we compute ROUGE over
SentencePiece tokens \cite{kudo-richardson-2018-sentencepiece} 
to avoid inconsistencies in tokenization among languages.

We measure the extent to which generated summaries are
faithful to their input using textual entailment
\cite{falke-etal-2019-ranking, kumar-talukdar-2020-nile, honovich-etal-2022-true-evaluating, whitehouse-etal-2023-webie, huot-etal-2024-mplan}.  Specifically, for our entailment classifier, we fine-tuned mT5-XXL \cite{xue-etal-2021-mt5} on two NLI datasets, namely ANLI \cite{nie-etal-2020-adversarial} and XNLI \cite{conneau-etal-2018-xnli}.  
Following previous work \cite{aharoni-etal-2023-multilingual,huot-etal-2024-mplan}, 
for each sentence in the summary, we compute its entailment probability given the input and report the average across sentences.

We also assess if a summary concisely represents the
information in the source article using a recently proposed metric
trained on the \textsc{seahorse} benchmark \cite{clark-etal-2023-seahorse}, which is a large-scale
collection of human ratings on various dimensions of system summary
quality across multiple languages, datasets, and models. We use a
publicly available mT5-XXL model \cite{xue-etal-2021-mt5} fine-tuned on 
binary conciseness judgments.\footnote{\url{https://huggingface.co/google/seahorse-large-q6}}

\begin{table}[t]
\centering
\scalebox{0.67}{
\begin{tabular}{@{}lr|ccc|ccc@{}}
\toprule
&  \multicolumn{1}{l|}{}
 & \multicolumn{3}{c|}{\sc \textbf{XLSum}}
 & \multicolumn{3}{c}{\sc  \textbf{XWikis}} \\
 & \multicolumn{1}{c|}{Params} &  R-L &  NLI &  SH & R-L &  NLI &  SH \\
\midrule

Reference & \multicolumn{1}{c|}{---} & --- & 48.50 & 31.65 & --- & 39.20 & 25.19 \\

\midrule

 Full FT &  100\%       
 & \bf 31.11 & 42.93 & 31.64
 & \bf 34.08 & 41.04 & 25.19 \\

 FT-Att & 20\% 
 & 30.88 & 50.32 & \bf 36.12
 & 32.22 & 37.06 & 24.20 \\
\midrule

{LoRA}-512 & 13.3\% 
 & 29.81 & 42.58 & 30.16
 & 33.38 & 40.48 & 24.78 \\

{LoRA}-64 & 1.7\% 
& 29.79 & 45.51 & 31.80
& 34.04 & 45.34 & 27.02 \\

{LoRA}-16 & 0.4\% 
 & 29.77 & 48.48 & 33.25
 & 33.80 & 46.10 & 27.42 \\

{LoRA}-4 & 0.1\% 
 & 29.03 & \bf 51.16 & 34.42
 & 32.92 & \bf 47.43 & \bf 27.72 \\
\bottomrule
\end{tabular}
}
\caption{Results on XLSum and XWikis with PaLM 2-XXS trained in the
  high-data regime: full fine-tuning on all layers (Full FT), on
  attention layers (FT-Att), and LoRA-* (with different ranks). Params
  denotes the proportion of trainable parameters.  Best ROUGE-L (R-L),
  NLI, and \textsc{seahorse} (SH) conciseness scores (area under the
  ROC curve)  are in bold.  Reference shows  NLI and SH scores on the reference/target summaries.}
 \label{tab:all-data}
\end{table}

\section{Results and Analysis}
\label{sec:results}

This section presents empirical results with LoRA on multilingual
summarization. We report comparisons to full fine-tuning, following
different data availability scenarios.

\subsection{High-data Regime}
\label{sec:full}
In the high-data regime, we use the complete training set, including
all languages in XLSum and XWikis. In \autoref{tab:all-data}, we
compare conventional full fine-tuning on all layers (Full FT), and a
more constrained setting that exclusively updates attention layers
(\mbox{FT-Att}) against LoRA variants where attention layers are tuned
with different ranks (\mbox{$r=\{$4, 16, 64, 512$\}$}).  We report
results with \mbox{PaLM 2-XXS} and select the best checkpoints based
on ROUGE-L throughout. All differences between best-performing models
(shown in bold) and comparison models are statistically significant
(across metrics) using paired bootstrap resampling ($\mbox{p <
  0.01}$).  We also report NLI and \textsc{seahorse} scores for the
reference summaries to gain a sense of the optimum value range for
these metrics.

In terms of summary relevance, perhaps unsurprisingly, conventional
fine-tuning on all layers achieves the best ROUGE-L scores for XLSum
and XWikis.  Updating attention layers only results in competitive
performance on \mbox{XLSum}, however, it delivers a drop of~1.86
ROUGE-L points on XWikis.  All LoRA variants, even those with high
ranks, update fewer parameters than constrained fine-tuning.  Despite
remarkable efficiency in parameter updates, LoRA with rank 4 lags
behind full fine-tuning (by 2.08 \mbox{ROUGE-L} points on \mbox{XLSum} and
1.16 on XWikis). In general, we observe that expanding the parameter
update space through higher ranks enhances summary relevance. For
XWikis, LoRA with rank~64 is very close to full fine-tuning. However,
for \mbox{XLSum} where language diversity and data imbalances are more
pronounced, all LoRA variants fall short of full fine-tuning by more
than 1~ROUGE\nobreakdash-L point.  In line with
\citet{chen-etal-2022-revisiting}, we observe that LoRA becomes more
sensitive to learning rates with higher ranks, requiring more careful
hyper-parameter tuning.

With regard to summary faithfulness and conciseness, we note that LoRA
achieves superior performance compared to full fine-tuning, with lower
rank settings exhibiting better NLI and \textsc{seahorse} scores.  We
further examined the length of the summaries obtained from full
fine-tuning and LoRA models. On XLSum, fully fine-tuned summaries are
on average 52.13 tokens long (using the SentencePiece tokenizer),
while LoRA summaries (rank~4) are slightly shorter with an average
length of~48.82. This difference in conciseness is further mirrored in
the \textsc{seahorse} scores for the two types of summaries.

Interestingly, compared to the reference, both full fine-tuning and
LoRA demonstrate higher conciseness. The predicted summaries also
overall show better NLI scores.  Example summaries are provided
\autoref{examples}, while additional results and language-specific
performance are included in \autoref{add_results}.

\paragraph{Takeaways} When training data is available, full
fine-tuning yields the most relevant and informative summaries. LoRA
is a competitive alternative, particularly when considering summary
faithfulness and conciseness. LoRA performance can be further enhanced
with higher ranks, although more careful hyper-parameter tuning is
generally required.

\begin{figure}[t!]
\centering
    \includegraphics[width=1\linewidth]{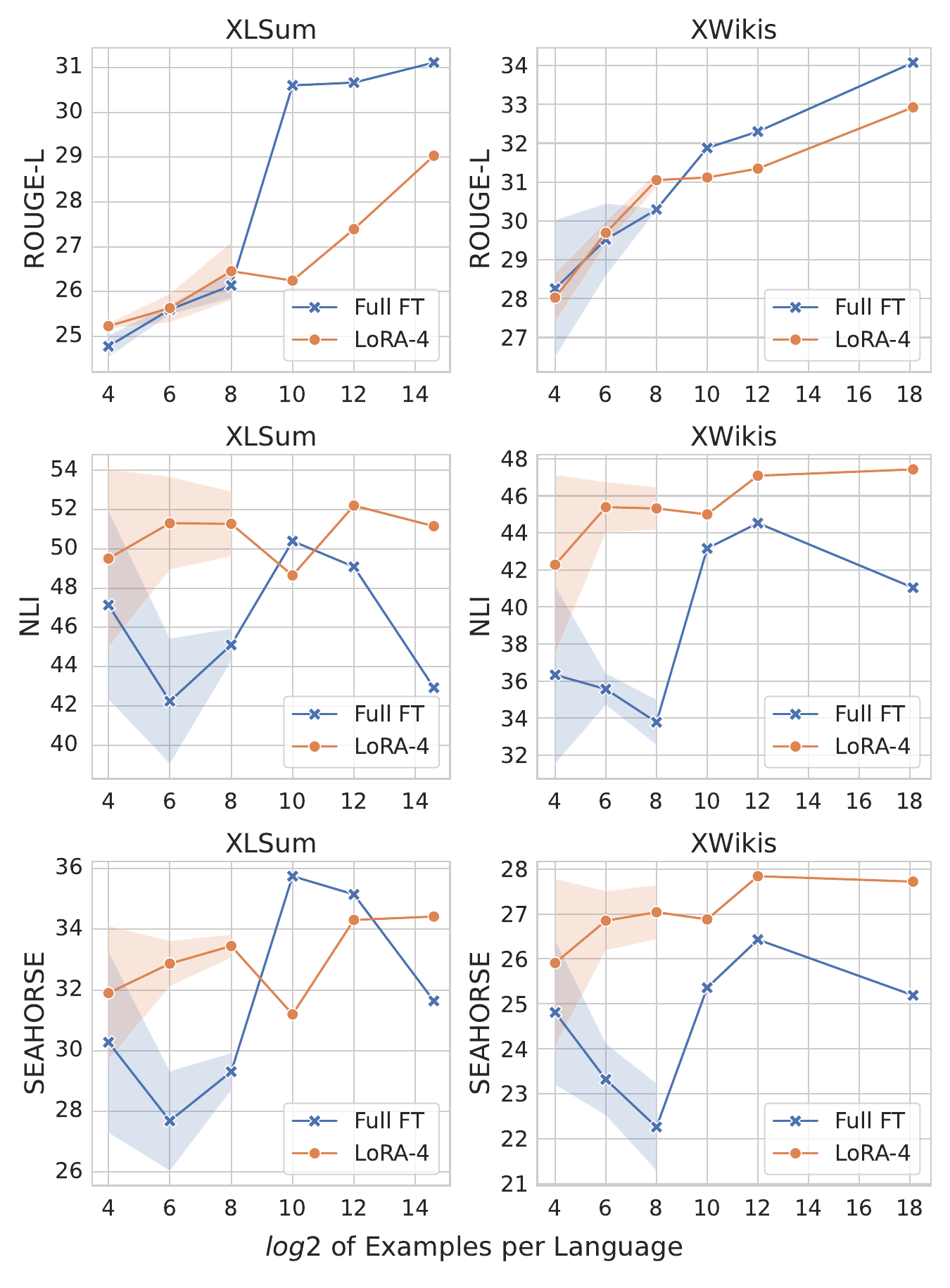}
    \caption{Results on the XLSum and XWikis datasets with PaLM 2-XXS
      trained in the low $\rightarrow$ high-data regime: Full FT
      vs. LoRA-4. Results for up to 256 examples per language are
      averaged over three seeds, with standard deviation shown in
       shaded areas.  }
\label{fig:lowdata}
\end{figure}

\subsection{Low-data Regime}
\label{sec:low}

We compare full fine-tuning against LoRA in the low-data regime
where limited training data is available. From this section onward, we
focus on LoRA with rank~4 and full fine-tuning on all
  layers.

We randomly sample 16, 64, and 256
training examples per language for both XLSum and XWikis.  To ensure the robustness of our results, we conduct experiments with three different
seeds, each with a unique set of samples. To examine how performance
evolves as we increase our training samples, 
we further present experiments with 1,024 and 4,096 examples per
language for both datasets.\footnote{When the number of training
samples is set to~4,096, three languages in XLSum already lack
sufficient data, so we refrain from selecting more examples per language.}
We set the number of validation samples to match that of the training
data.  As before, we select the best checkpoint based on ROUGE-L and
subsequently evaluate on the entire test set.

\autoref{fig:lowdata} shows the performance of PaLM~2-XXS with full fine-tuning and LoRA, when the number of training
examples per language varies from 16 to the entire dataset.  The
x-axis corresponds to the number of examples per language on \(log\) scale; the
high-data setting is approximated by~\mytilde$2^{14.6}$ examples in
XLSum and \mytilde$2^{18.1}$ in XWikis.
For training data with 256 or fewer samples, we show the standard deviation
with shaded areas.  We observe that LoRA achieves overall better faithfulness (NLI) and conciseness (\textsc{seahorse}) than full fine-tuning.  For ROUGE-L, LoRA demonstrates
advantages in low-data scenarios, while full fine-tuning delivers a
performance boost when increasing the number of examples from 256
to~1,024.  

In addition, full fine-tuning is sensitive to checkpoint selection in
the low-data regime, due to its susceptibility to overfitting,
requiring more frequent validation.  In comparison, the training
process for LoRA is more stable.

\paragraph{Takeaways} In low-data scenarios, LoRA is a better
alternative to full fine-tuning.  LoRA delivers consistently competitive
or even superior results with the additional advantage of efficient and
stable training.

\subsection{Cross-lingual Transfer}
We now focus on cross-lingual transfer in multilingual summarization
and explore two common scenarios, namely zero- and few-shot learning. 
For LoRA, we focus on rank 4 in all experiments.

\paragraph{Zero-shot Transfer from English}
\begin{table}[!t]
\centering
\scalebox{0.66}{
\begin{tabular}{@{}ll|ccc|ccc@{}}
\toprule

& \multicolumn{1}{l|}{Test}
 & \multicolumn{3}{c|}{\sc \textbf{XLSum}} & \multicolumn{3}{c}{\sc  \textbf{XWikis}} \\
 & Languages &  R-L &  NLI &  SH &  R-L &  NLI &  SH \\

\midrule
 Full FT & \multirow{2}{*}{Non-English}
 & ~~5.20 & ~~4.49 & ~~6.88 & 17.51 & 35.95 & 22.43 \\
 LoRA-4 &
 &  \bf 21.13 & \bf 39.07 & \bf 23.08 & \bf 23.86 & \bf 45.54 & \bf 25.96 \\
\midrule

Full FT & \multirow{2}{*}{English}
 & \bf 32.58 & 57.09 & 38.01 & \bf 36.59 & \bf 53.59 & \bf 30.81 \\
 LoRA-4 &
 &  32.21 & \bf 63.13 & \bf 43.44 &  34.07 & 49.94 & 29.01 \\

\bottomrule
\end{tabular}}
\caption{Zero-shot cross-lingual transfer using full fine-tuning (Full FT)
  and LoRA (rank 4); PaLM 2-XXS models are trained and validated on
  English and tested on all other languages (Non-English) and English
  only.  Best
ROUGE-L (R-L), NLI, and \textsc{seahorse} (SH) conciseness scores (area under the ROC curve) are in bold.}
\label{tab:zero-en}
\end{table}
 We first consider a typical scenario where only
English training data is available, i.e., training and
validation are carried out using English examples, whilst the model is
tested on new languages.

\autoref{tab:zero-en} shows the performance of \mbox{PaLM 2-XXS}
with full fine-tuning and LoRA. We separate results on English as they
are not zero-shot (second block in \autoref{tab:zero-en}) and they broadly align with our findings in
Section~\ref{sec:full} (high-data regime). Full fine-tuning generally
outperforms LoRA except for NLI and SH on English XLSum.  In the
cross-lingual transfer scenario (first block in
\autoref{tab:zero-en}), full fine-tuning performs exceptionally poorly
across metrics and languages on XLSum.  The gap is smaller for XWikis
as only four non-English languages are covered and all but Chinese are
in the Indo-European family. Further examination of the model output shows that the generated text is mostly in English rather than the target language.
The model appears to comprehend the new language (i.e.,~input
documents), however, it struggles to generate output accordingly.


\begin{figure}[t!]
\centering
    \includegraphics[width=1\linewidth]{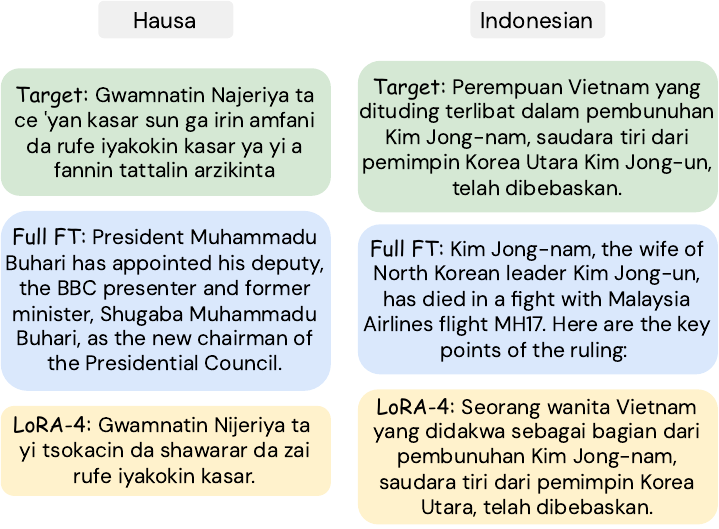}
    \caption{XLSum output examples: zero-shot transfer from English
      using Full FT and LoRA with PaLM \mbox{2-XXS}. Full FT fails to
      generate summaries in the target language and the content is
      off-topic.}
\label{fig:examples}
\end{figure}

\autoref{fig:examples} illustrates XLSum examples of model output for  Hausa and Indonesian. In both cases, Full FT summaries are in English, and off-topic (the Hausa article discusses the Nigerian government's decision to close its borders, while the Indonesian one reports on the murder of Kim Jong-un). 
In \autoref{add_results}, we provide per-language results which
highlight that for zero-shot transfer from English, full fine-tuning
consistently lags behind LoRA in \textit{every} language, even in
cases where languages are well-represented in the pre-training phase
of PaLM~2 or are considered linguistically close to
English.\footnote{See Table~21 in \citet{anil2023palm} regarding
the distribution of languages used in the pre-training of PaLM 2.}
This catastrophic forgetting behavior echoes the findings in
\citet{vu-etal-2022-overcoming}.

\paragraph{Zero-shot Transfer from Multiple Languages}
\label{sec:multi}
We  extend our study of zero-shot cross-lingual transfer to
scenarios where training data is available in multiple languages
rather than just English. 

For XLSum, we create a training data pool of 10 languages from eight
distinct linguistic families, each with substantial training data.
These languages include Arabic (\textsc{ar}), (Simplified) Chinese
(\textsc{zh}), English (\textsc{en}), Hausa (\textsc{ha}), Hindi
(\textsc{hi}), Indonesian (\textsc{id}), Persian (\textsc{fa}),
Portuguese (\textsc{pt}), Swahili (\textsc{sw}), and Turkish
(\textsc{tr}). Additionally, we select 10 test languages: Azerbaijani
(\textsc{az}), Bengali (\textsc{bn}), Japanese (\textsc{ja}), Kirundi
(\textsc{rn}), Korean (\textsc{ko}), Nepali (\textsc{ne}), Scottish
Gaelic (\textsc{gd}), Somali (\textsc{so}), Thai (\textsc{th}), and
Yoruba (\textsc{yo}). Test languages are selected so that they are
maximally diverse, each representing a unique language
family.\footnote{See \autoref{sec:xlsum} for details of language
families in XLSum.} For XWikis, we adopt a \textit{leave-one-out}
approach, since it only covers five languages. We rotate through the
available languages, one for testing and four for training.

In addition to full fine-tuning and LoRA, we report experiments with
language-specific LoRA modules, each trained on examples from one
language.  An advantage of such specialized modules is their
scalability and adaptability.  When additional languages become
available, there is no need to re-train the entire model; it is
sufficient to add a new language-specific module.  During inference,
we can also flexibly experiment with various LoRA modules or weight
composition methods.  As mentioned in
Section~\ref{sec:weight_composition}, weight composition is an active
research area that has demonstrated effectiveness across a spectrum of
applications.  We adopt a simple approach that computes the weighted
average of all available modules.

\begin{figure}[t!]
    \centering
    \begin{subfigure}{0.5\textwidth}
        \centering
       \hspace*{-.2cm}\includegraphics[width=.95\linewidth]{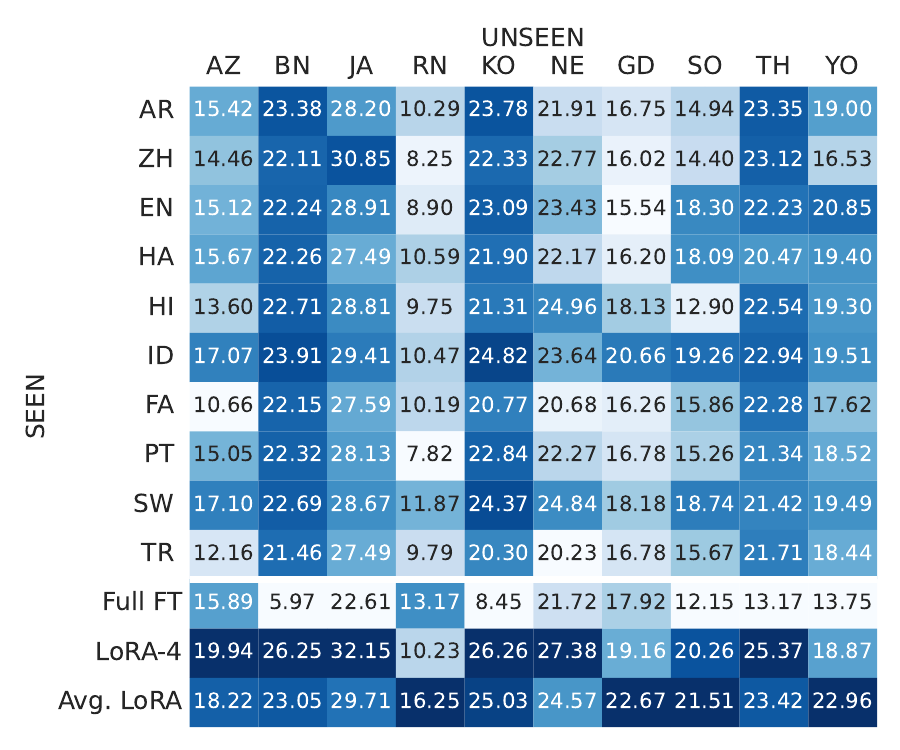}
        \vspace*{-1mm}
        \caption{ROUGE-L scores for 10~test languages on XLSum.}
        \label{fig:top}
    \end{subfigure}
   \vspace*{2mm}
    \begin{subfigure}{0.5\textwidth}
        \centering
\hspace*{-.2cm}\includegraphics[width=0.8\linewidth]{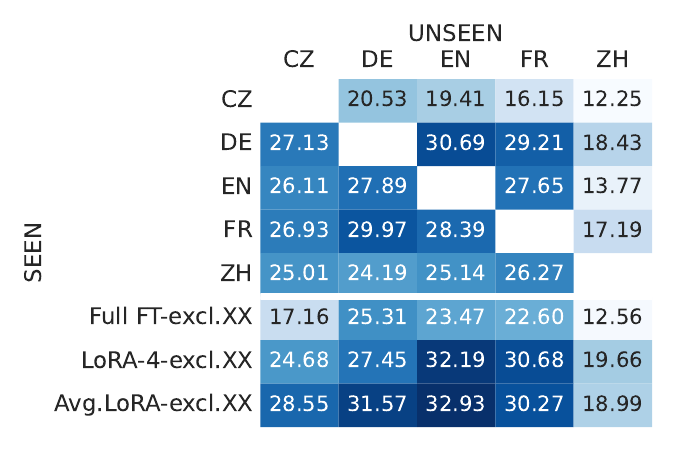}
        \vspace*{-1mm}
        \caption{ROUGE-L scores for five languages on XWikis.} %
        \label{fig:bottom}
    \end{subfigure}
\caption{Zero-shot cross-lingual transfer on XLSum
  (top) and XWikis (bottom); PaLM 2-XXS models are trained on one language
  (\texttt{SEEN})  and tested on another (\texttt{UNSEEN}). We
  also show results with full fine-tuning on \emph{all} seen languages
  (Full FT), LoRA, and (average) weighted combination of
  language-specific LoRA modules (\texttt{Avg.LoRA}); \texttt{excl.XX}
  in XWikis denotes \textit{leave-one-out} training, excluding the
  test language.}
    \label{fig:heatmap}
\end{figure}

\autoref{fig:heatmap} shows the heatmaps of the ROUGE\nobreakdash-L scores
for models trained in one language and tested on another. Rows represent source models from
\texttt{SEEN} languages and columns represent \texttt{UNSEEN} test languages. The color scale is column-wise normalized to
provide a comparative view of the performance of the best and
worst models for each \texttt{UNSEEN} test language. In the bottom three rows, we also
illustrate the performance of models trained on multiple seen
languages and tested on unseen ones. We experiment with full
fine-tuning, LoRA (rank 4), and weighted average LoRA.

We observe from \autoref{fig:heatmap} that: (i)~full fine-tuning
consistently lags behind LoRA in zero-shot cross-lingual transfer,
even with a diverse collection of languages besides English; (ii) the
weighted average of language-specific LoRA modules (\texttt{Avg.LoRA})
and LoRA (trained on all available languages together) benefit
different unseen languages.  Particularly for XLSum, lower-resource
languages (i.e., RN, GD, SO, and YO), exhibit superior performance
with language-specific LoRA training.  We hypothesize that in the
default LoRA setting (i.e.,~joint training across languages),
high-resource languages hinder the effective learning of 
low-resource languages; and~(iii)~languages with similarities
  demonstrate better transferability, as exemplified by transferring
  ZH to JA and SW to RN on XLSum, and the Indo-European languages on
  XWikis.

\paragraph{Few-shot Cross-lingual Transfer}

Finally, we consider scenarios where some examples are available in
the target languages and explore effective strategies for utilizing
them. We follow the previous section and assume that models
have been already trained on (seen) languages with sufficient data.
One approach is to \textit{continue} training these models using
target language examples. Therefore, if the starting checkpoint was obtained
from full fine-tuning on seen languages, we continue with full
fine-tuning on the new languages. We also adopt the same strategy for
LoRA.

Another widely-used technique is \textit{in-context learning}, where
input and output examples are concatenated to form in-context
demonstrations. Despite promising results in many LLM applications
\cite{NEURIPS2020_1457c0d6, wei2022chain}, in-context learning becomes
less practical in the domain of multilingual summarization where
models are expected to process long articles, which is
memory-intensive, especially as the number of examples grows. Instead,
we experiment with the recently proposed few-shot LoraHub learning
approach (Section~\ref{sec:lora}).  The original formulation of
LoraHub \cite{huang2023LoraHub} does not assume any prior knowledge of
the available LoRA modules which are randomly sampled and initialized
with zero weights (i.e.,~starting from a general-purpose pre-trained
LLM). We initialize LoraHub with the weighted sum of $N$
language-specific LoRA modules and assign a weight of $1/N$ to each
module. The composition of modules fine-tuned on the same task, albeit
in different languages, offers a stronger baseline compared to a
pre-trained LLM.

\begin{table}[t]
\centering
\scalebox{0.65}{
\begin{tabular}{ll|ccc|ccc}
\toprule

& \multicolumn{1}{c|}{}& \multicolumn{3}{c|}{\sc \textbf{XLSum}} & \multicolumn{3}{c}{\sc \textbf{XWikis}} \\
 &
 &  R-L &  NLI &  SH &  R-L &  NLI & SH \\

\midrule
\multirow{3}{2.5em}{\sc {zero-shot}}
 & Full FT & 14.48 & 28.87 & 13.71 & 20.22 & 30.17 & 16.26 \\
 & LoRA-4 & 22.59 & 37.39 & 24.21 & \bf 28.46 & 48.31 & 26.40 \\
 & Avg. LoRA & \bf 22.74 & \bf 49.14 & \bf 32.44 & 26.93 & \bf 49.29 & \bf 26.86 \\
\midrule
    
\multirow{3}{2em}{\sc {16-shot}}
 & Full FT (\textsc{cl}) & 22.31 & 30.15 & 18.79 & 26.90 & 34.17 & 21.82 \\
 & LoRA-4 (\textsc{cl}) & \bf 24.71 & \bf 41.12 & \bf 26.47 & \bf 30.05 & 45.90 & \bf 28.20 \\
 & LoraHub & 23.37 & 38.95 & 26.07 & 27.59 & \bf 47.45 & 25.84 \\
\midrule
    
\multirow{3}{2em}{\sc {64-shot}}
 & Full FT (\textsc{cl}) & 24.30 & 30.65 & 19.57 & 28.73 & 39.42 & 24.16 \\
 & LoRA-4 (\textsc{cl}) & \bf 25.94 & \bf 42.07 & 27.66 & \bf 31.08 & 45.12 & \bf 28.05 \\
 & LoraHub & 24.21 & 41.34 & \bf 28.02 & 27.66 & \bf 48.09 & 26.56 \\

\bottomrule
\end{tabular}
}
\caption{Cross-lingual transfer on 10 XLSum languages and five XWikis
  languages (using \textit{leave-one-out} training) for PaLM 2-XXS model.  16-
  and 64-shot experiments show average results from three different
  seed runs. For \emph{continued learning} (\textsc{cl}), we use a 14/2 and
  60/4 training/validation split.  Best 
  results 
  are in bold. Results for individual languages are in
  Tables~\ref{tab:few-shot-xlsum} and~\ref{tab:few-shot-xwikis} in~\autoref{add_results}.  }
 \label{tab:few-shot}
\end{table}

We consider two few-shot settings, with 16 or 64 target language
examples, simulating practical scenarios where human annotators or
experts create a few examples for low-resource languages.  We compare
few-shot \textit{continued learning} and \textit{LoraHub learning},
using the same examples. To ensure robustness, all experiments are conducted
on three different sets of examples, and we report the average. For
continued learning, we split the examples into training and
validation using 14/2 and 60/4 splits.  For LoraHub, we use the
\texttt{Nevergrad}
toolkit\footnote{\url{https://facebookresearch.github.io/nevergrad}}
for black-box optimization. We empirically compared ROUGE-L and loss as performance metrics guiding the optimization and
found that ROUGE-L led to more stable results.

\autoref{tab:few-shot} presents our results on XLSum and XWikis with
16- and 64-shots, averaged across test languages. Zero-shot results
are also included for comparison.  Our analysis supports the following
observations: (i) with only a few target language examples (e.g.,~16),
full fine-tuning sees a remarkable improvement, resulting in an
average boost of~7.8 ROUGE-L points on XLSum and 6.7~on XWikis,
corroborating the findings of \citet{lauscher-etal-2020-zero}; (ii)
LoraHub slightly enhances ROUGE-L performance compared to (zero-shot)
weighted-average on XLSum with only~16 examples; (iii) LoRA continued learning
consistently outperforms full fine-tuning and LoRAHub in terms of
ROUGE-L and SH; however, LoRAHub is superior in terms of NLI for XWikis.

\begin{table}[!t]
\centering
\scalebox{0.7}{
\begin{tabular}{lr|ccc|ccc}
\toprule
 & 
\multicolumn{1}{c|}{} & \multicolumn{3}{c|}{\sc \textbf{XLSum}}
 & \multicolumn{3}{c}{\sc \textbf{XWikis}} \\
 & Params &  R-L &  NLI &  SH &  R-L &  NLI &  SH \\
\midrule

 Full FT & 100\% & \bf 36.99 & 58.72 & 41.92 & \bf 39.65 & 46.03 & 28.01 \\
 LoRA-4 & 0.04\% & 36.29 & \bf 61.64 & \bf 43.99 & 39.25 & \bf 47.56 & \bf 28.30 \\
\bottomrule

\end{tabular}}
\caption{Results on XLSum and XWikis datasets with PaLM 2-S trained in
  the high-data regime: Full FT and LoRA (rank 4).
  Params denotes the proportion of trainable parameters. Best results
  are in bold.}
\label{tab:palm2s}
\end{table}

\paragraph{Takeaways}
In cross-lingual transfer scenarios, LoRA achieves consistently
superior performance compared to full fine-tuning.  LoRA continued learning shows particular promise when only a small number of examples are available in the target language.

 \begin{table}[!t]
\centering
\scalebox{0.65}{
\begin{tabular}{ll|ccc|ccc}
\toprule

 & \multicolumn{1}{l|}{Test}
 & \multicolumn{3}{c|}{\sc \textbf{XLSum}} & \multicolumn{3}{c}{\sc \textbf{XWikis}} \\
 & Languages &  R-L &  NLI &  SH &  R-L &  NLI &  SH \\

\midrule
 Full FT & \multirow{2}{*}{Non-English}
 & 33.22 & 60.72 & 41.96 & 35.70 & 46.27 & 27.51 \\
 LoRA-4 &
 &  \bf 33.31 & \bf 64.18 & \bf 43.98 & \bf 36.00 & \bf 47.23 & \bf 28.69 \\
\midrule

 Full FT & \multirow{2}{*}{English} 
 & \bf 40.38 & 71.21 & 45.82 & \bf 42.03 & \bf 51.76 & 28.95 \\
 LoRA-4 &
 & 39.61 & \bf 78.05 & \bf 47.02 & 41.53 & 50.09 & \bf 29.07 \\

\bottomrule

\end{tabular}}
\caption{Zero-shot transfer on XLSum and XWikis using
  Full FT and LoRA (rank~4). PaLM 2-S models are trained
  and validated on English and tested on all other languages
  (Non-English) and English only. Best results
  are in bold.  }
\label{tab:palm2s_crosslingual}
\end{table}

\section{Scaling Up}
We extend our analysis to the larger PaLM 2-S model, focusing on the
high-data regime and zero-shot cross-lingual transfer using English
data.  Our results are summarized in \autoref{tab:palm2s}
and \autoref{tab:palm2s_crosslingual}.

Interestingly, LoRA and full fine-tuning achieve 
similar performance, with  LoRA taking the lead in cross-lingual transfer (see first block
in \autoref{tab:palm2s_crosslingual}).  We hypothesize that when
using the larger \mbox{PaLM 2-S} model, the increased capacity makes up for
the small percentage of trainable parameters in LoRA (only 0.04\% of
the parameters), allowing it to
benefit more from high-data regime training.  At the same time, the larger model is more robust and does not exhibit catastrophic
forgetting during full fine-tuning.  As a result, we see that full
fine-tuning performs on par with LoRA in the zero-shot cross-lingual
setting (see \autoref{tab:palm2s_crosslingual}).

\paragraph{Takeaways}  For larger models such as \mbox{PaLM 2-S}, LoRA achieves on-par performance with full fine-tuning but is a better choice
when considering computational efficiency.

\section{Conclusions}
\label{sec:conclusions}

In this paper, we explored the effectiveness of LoRA on multilingual
summarization across a diverse range of scenarios primarily determined
by the availability of training data. We summarize our key findings by
comparing the computationally efficient LoRA against full fine-tuning.

LoRA achieves \textbf{superior performance} to full fine-tuning in zero-shot and few-shot cross-lingual transfer
scenarios, and \mbox{low-data} settings (e.g.,
training data with fewer than~1K samples). This is most pronounced with
smaller models (e.g.,~PaLM \mbox{2-XXS}).  In the specific case of few-shot learning, LoRA continued learning outperforms LoraHub.
LoRA also achieves overall superior summary faithfulness and conciseness
across various
scenarios.

For larger models like PaLM 2-S, LoRA exhibits \textbf{on-par
  performance} to full fine-tuning. This suggests that model capacity matters. Notably, for smaller
models like PaLM 2-XXS, LoRA displays \textbf{worse performance} in the 
full fine-tuning (high-data) regime, when said performance is measured
via ROUGE-L, but is consistently superior in terms of faithfulness
and conciseness.

Taken together, our results underscore the utility of
PEFT methods for complex multilingual tasks and
cross-lingual transfer. Avenues for future work include few-shot
transfer and effective ways to combine LoRA modules, e.g.,~by
learning which modules to activate for different
tasks or languages \cite{ponti-etal-2023-combining, lin2024multitask}. It would also be
interesting to reproduce our results across varied LLMs and broader
multilingual generation tasks, beyond summarization.

\section*{Limitations}

We identify the following limitations of our work:
\begin{itemize}
    \item We focused exclusively on decoder-only models.
Future work could explore a wider range of LLMs, including
encoder-decoder models.  We anticipate the observations gained from
decoder-only models to largely align with those from encoder-decoder
models, thus generalizing our findings.

\item In our cross-lingual transfer
studies, we only considered LoRA models with a rank of 4, due to
computational considerations.  Expanding to additional LoRA settings
would allow us to perform a more thorough comparison. 

\item Our
experiments have exclusively focused on multilingual summarization
tasks. Extending our study to a wider range of multilingual text
generation tasks with long input and output would provide a more
comprehensive perspective on the capabilities and limitations of LoRA.

\item We concentrate on LoRA as a representative parameter-efficient fine-tuning approach, however, extending our study to other PEFT methods could bring more insights.
\end{itemize}

\bibliography{custom}
\bibliographystyle{acl_natbib}

\appendix

 \section{Datasets}
 \label{sec:xlsum}
 
 \autoref{tab:xlsum-data} and \autoref{tab:xwikis-data} show the language families and the number
of training examples per language in the XLSum and XWikis datasets.

\section{TPU Computational Requirements}
 \label{sec:tpu}
 With regard to computational requirements, memory-intensive
 experiments were conducted on Cloud TPU v4, while less
 memory-intensive ones were run on Cloud TPU v3. For instance, full
 fine-tuning jobs in the high-data regime required up to 64 TPUs v4,
 while less memory-intensive jobs such as LoRA with continued learning
 (for cross-lingual transfer) required 8 TPUs v3.

 \section{Additional Results}
 \label{add_results}
 \autoref{tab:all-rouge} shows  ROUGE-1, ROUGE-2, and ROUGE\nobreakdash-L scores for LoRA and full fine-tuning with PaLM~2-XXS on the two datasets.
 We additionally report activating LoRA tuning on Feed Forward layers with different ranks.

\begin{table}[ht]
\centering

\addtolength{\tabcolsep}{-3pt}
\scalebox{0.85}{
\begin{tabular}{lclr}
\toprule
\textbf{Language} & \textbf{ISO} & \textbf{Language Family} & \textbf{\# Train} \\
\midrule
English             & EN & Indo-European     & 306,522 \\
Hindi               & HI & Indo-European     & 70,778  \\
Urdu                & UR & Indo-European     & 67,665  \\
Russian             & RU & Indo-European     & 62,243  \\
Portuguese          & PT & Romance           & 57,402  \\
Persian             & FA & Indo-Iranian      & 47,251  \\
Ukrainian           & UK & Slavic            & 43,201  \\
Indonesian          & ID & Austronesian      & 38,242  \\
Spanish             & ES & Romance           & 38,110  \\
Arabic              & AR & Semitic           & 37,519  \\
Chinese-Traditional & ZH & Sino-Tibetan      & 37,373  \\
Chinese-Simplified  & ZH & Sino-Tibetan      & 37,362  \\
Vietnamese          & VI & Austroasiatic     & 32,111  \\
Turkish             & TR & Turkic            & 27,176  \\
Tamil               & TA & Dravidian         & 16,222  \\
Pashto              & PS & Indo-Iranian      & 14,353  \\
Marathi             & MR & Indo-Aryan        & 10,903  \\
Telugu              & TE & Dravidian         & 10,421  \\
Welsh               & CY & Celtic            & 9,732   \\
Pidgin              & PI & Unknown           & 9,208   \\
Gujarati            & GU & Indo-European     & 9,119   \\
French              & FR & Romance           & 8,697   \\
Punjabi             & PA & Indo-Iranian      & 8,215   \\
Bengali             & BN & Indo-European     & 8,102   \\
Swahili             & SW & Bantu             & 7,898   \\
Serbian-Latin       & SR & Indo-European     & 7,276   \\
Serbian-Cyrillic    & SR & Indo-European     & 7,275   \\
Japanese            & JA & Japonic           & 7,113   \\
Thai                & TH & Kra-Dai Languages & 6,616   \\
Azerbaijani         & AZ & Turkic            & 6,478   \\
Hausa               & HA & Afro-Asiatic      & 6,418   \\
Yoruba              & YO & Niger-Congo       & 6,350   \\
Oromo               & OM & Afro-Asiatic      & 6,063   \\
Somali              & SO & Afro-Asiatic      & 5,962   \\
Nepali              & NE & Indo-Aryan        & 5,808   \\
Amharic             & AM & Semitic           & 5,761   \\
Kirundi             & RN & Bantu             & 5,746   \\
Tigrinya            & TI & Semitic           & 5,451   \\
Uzbek               & UZ & Turkic            & 4,728   \\
Burmese             & MY & Sino-Tibetan      & 4,569   \\
Korean              & KO & Koreanic          & 4,407   \\
Igbo                & IG & Niger-Congo       & 4,183   \\
Sinhala             & SI & Indo-European     & 3,249   \\
Kyrgyz              & KY & Turkic            & 2,266   \\
Scottish-Gaelic     & GD & Celtic            & 1,313  
 \\

\bottomrule
\end{tabular}}
\caption{Language family and number of training examples per language
  in the XLSum dataset.
}
\label{tab:xlsum-data}
\end{table}

 \begin{table}[t]
\centering

\addtolength{\tabcolsep}{0pt}
\scalebox{0.9}{
\begin{tabular}{lclr}
\toprule
\textbf{Language} & \textbf{ISO} & \textbf{Language Family} & \textbf{\# Train} \\
\midrule
English             & EN & Indo-European     & 624,178 \\
German              & DE & Indo-European     & 390,203  \\
French              & FR & Indo-European     & 323,915  \\
Czech               & CS & Indo-European     & 61,224  \\
Chinese             & ZH & Sino-Tibetan      & 31,281 \\

\bottomrule
\end{tabular}}
\caption{Language family and  number of training examples per language
  in the XWikis dataset.
}
\label{tab:xwikis-data}
\end{table}

    \begin{table*}[t]
\centering
\scalebox{0.8}{
\addtolength{\tabcolsep}{0pt}
\begin{tabular}{lllc|cccccc}
\toprule
\multirow{2}{*}{\textbf{PaLM 2-XXS}}
& \multicolumn{1}{l}{\multirow{2}{*}{\parbox{0.2\textwidth}{Trainable Layers}}}      &  & 
\multicolumn{1}{c|}{\multirow{2}{*}{ \parbox{0.1\textwidth}{Params}}} 
 &\multicolumn{3}{c}{\sc \textbf{XLSum}}       & \multicolumn{3}{c}{\sc \textbf{XWikis}}   \\
 &  & &    & R-L & R-1 & R-2
& {R-L} & {R-1} & {R-2}
\\
\cmidrule(r){1-4} \cmidrule(lr){5-7} \cmidrule(l){8-10} 
\multicolumn{1}{l}{\multirow{2}{*}{\textbf{Full FT}}} & All Layers & & ~100\%       
 & 31.11  & 41.66           & 21.78    
 & 34.08  & 42.68           & 24.37     
 \\
& Attention Layers & & ~~~20\% 
& 30.88   & 41.17           & 21.43     
& 32.22           & 41.48           & 22.54  
\\  

\cmidrule{1-10} 
\multirow{7}{*}{\textbf{LoRA}} &
\multicolumn{1}{l}{\multirow{4}{*}{\parbox{0.2\textwidth}{Attention Layers}}}  &  \textit{rank 512} & 13.3\% 
&   29.81 &	40.33&	20.25
&	33.38&	41.52&	23.63  
\\
&  & \textit{rank 64} & ~1.7\% 
& 29.79           & 39.98           & 20.18      
& 34.04           & 41.58           & 24.28 
\\
&  & \textit{rank 16} & ~0.4\% 
 & 29.77           & 39.75           & 20.09     
 & 33.80           & 41.34           & 24.14     

 \\
& & \textit{rank 4}  & ~0.1\% 
& 29.03           & 38.83           & 19.28        
& 32.92           & 39.97           & 23.27       
\\

\cmidrule{2-10}
& \multicolumn{1}{l}{\multirow{3}{*}{\parbox{0.2\textwidth}{Attention + FFN\\Layers}}} & \textit{rank 64} & 5.4\%     
 & 29.45           & 39.64           & 19.79           & 33.59           & 41.37           & 23.79              \\
& & \textit{rank 16} & 1.4\%   
& 29.79           & 39.99           & 20.17           & 33.55           & 41.11           & 23.95                \\
&& \textit{rank 4}  & ~0.3\% 
 & 29.67           & 39.76           & 20.02    
 & 33.70           & 40.82           & 24.05     
 \\

\bottomrule
\end{tabular}
}
\caption{Results on XLSum and XWikis datasets with
 PaLM 2-XXS trained in the high-data regime: full fine-tuning on all
 layers, full fine-tuning on attention layers, and LoRA (with
 different ranks). Params denotes the proportion of trainable
 parameters.}
 \label{tab:all-rouge}
\end{table*}

   \autoref{tab:few-shot-xlsum}, \autoref{tab:few-shot-xlsum-nli}, and \autoref{tab:few-shot-xlsum-sh}
   show ROUGE\nobreakdash-L, NLI, and \textsc{seahorse} few-shot learning results for individual languages
   on XLSum. \autoref{tab:few-shot-xwikis}, \autoref{tab:few-shot-xwikis-nli}, and \autoref{tab:few-shot-xwikis-sh}
   show ROUGE-L, NLI, and \textsc{seahorse} few-shot learning results for individual languages
   on XWikis.

\begin{table*}[t]
\centering
\scalebox{0.8}{
\addtolength{\tabcolsep}{-2pt}
\begin{tabular}{ll|ccccccccccc}
\toprule
    
&\multicolumn{1}{l|}{\textbf{PaLM 2-XXS}}
& {\textbf{AVG}} & {{AZ}} & {{BN}} & {{JA}}& {{RN}} & {{KO}} & {{NE}} & {{GD}} & {{SO}} & {{TH}} & {{YO}} \\
 \midrule

{\multirow{3}{*}{\parbox{0.07\textwidth}{\sc {zero-\\shot}}}}

& Full FT & 14.48 & 15.89          & ~~5.97           & 22.61          & 13.17          & ~~8.45           & 21.72          & 17.92          & 12.15          & 13.17          & 13.75   \\
  & LoRA-4  & 22.59 & \textbf{19.94} & \textbf{26.25} & \textbf{32.15} & 10.23          & \textbf{26.26} & \textbf{27.38} & 19.16          & 20.26          & \textbf{25.37} & 18.87  \\
    & Avg. LoRA & 

     \textbf{22.74}	& 18.22&	23.05&	29.71 &	 \textbf{16.25}	& 25.03 &	24.57 &	 \textbf{22.67}&	 \textbf{21.51}&	23.42&	 \textbf{22.96}
    \\
  \midrule
    
{\multirow{3}{*}{\parbox{0.07\textwidth}{\sc {16-\\shot}}}} &
    
 Full FT + \textit{continued learning} &22.31&	16.64&	22.95&	28.28&17.02	&24.31&	26.94&	19.56	&19.66&	23.58	&24.18
    \\       

  & LoRA-4 + \textit{continued learning}     & \textbf{24.71} &	\textbf{20.74} &	\textbf{26.19} &	\textbf{32.26} &	\textbf{17.82} &	\textbf{27.13} &	\textbf{27.82} &	23.00 &		22.26 &		24.30 &	\textbf{25.55}
  \\
  & LoraHub &
23.37 & 18.58 & 24.81 & 27.69 & 16.65 & 25.82 & 25.40 & \bf 24.83 & \bf 23.13 & \bf 24.71 & 22.05     \\
  
     \midrule
    
{\multirow{3}{*}{\parbox{0.07\textwidth}{\sc {64-\\shot}}}} &
    
  Full FT + \textit{continued learning}  & 24.30&	17.86&	22.80&	32.49&	\textbf{19.28}&	27.09&	28.89&	21.72&	22.17&	23.90&	26.83 
    \\
 & LoRA-4 + \textit{continued learning}   &	\textbf{25.94} &	\textbf{20.91} &	\textbf{26.08} &	\textbf{33.10} &	19.09 &	\textbf{28.43} &	\textbf{29.38} &	\textbf{25.78} &	\textbf{23.06} &	\textbf{25.48} &	\textbf{28.06}
  \\
  & LoraHub
  & 24.21 & 20.10 & 25.16 & 29.03 & 17.61 & 27.46 & 26.82 & 24.94 & 23.04 & 24.37 & 23.53         \\

\bottomrule
\end{tabular}
}
\caption{Cross-lingual transfer results (ROUGE-L) on 10 XLSum
  languages for PaLM 2-XXS model. 16- and 64-shot experiments
  show average results from three different seed runs. For \textit{continued learning}, we use a 14/2 and 60/4 split for
  training/validation.}
 \label{tab:few-shot-xlsum}
\end{table*} 

\begin{table*}[t]
\centering
\scalebox{0.8}{
\addtolength{\tabcolsep}{-2pt}
\begin{tabular}{ll|ccccccccccc}
\toprule
    
&\multicolumn{1}{l|}{\textbf{PaLM 2-XXS}}
& {\textbf{AVG}} & {{AZ}} & {{BN}} & {{JA}}& {{RN}} & {{KO}} & {{NE}} & {{GD}} & {{SO}} & {{TH}} & {{YO}} \\
 \midrule

{\multirow{3}{*}{\parbox{0.07\textwidth}{\sc {zero-\\shot}}}}

& Full FT & 28.87 & 19.17 & 28.28 & 35.27 & 24.00 & 30.76 & 46.44 & \bf 38.22 & 15.23 & 33.02 & 18.26 \\
  & LoRA  & 37.39 & 37.92 & 52.61 & 62.54 & 9.62 & 54.57 & 47.35 & 16.86 & 21.92 & 53.23 & 17.26  \\
    & Avg. LoRA & \bf 49.14 & \bf 45.54 & \bf 60.55 & \bf 66.37 & \bf 39.63 & \bf 66.44 & \bf 55.86 & 33.43 & \bf 34.26 & \bf 60.86 & \bf 28.47
    \\
  \midrule
    
{\multirow{3}{*}{\parbox{0.07\textwidth}{\sc {16-\\shot}}}} &
    
 Full FT + \textit{continued learning} & 30.15 & 15.83 & 46.29 & 37.55 & 17.55 & 35.61 & 42.22 & 20.07 & 26.69 & 39.74 & 20.00
    \\       

  & LoRA + \textit{continued learning}     & \bf 41.12 & 37.33 & \bf 56.45 & \bf 52.31 & 30.62 & \bf 58.08 & \bf 49.10 & 24.32 & \bf 33.37 & 45.02 & \bf 24.58
  \\
  & LoraHub &
38.95 & \bf 37.76 & 47.95 & 48.17 & \bf 35.52 & 49.29 & 40.90 & \bf 32.14 & 26.29 & \bf 52.08 & 19.40 \\
  
     \midrule
    
{\multirow{3}{*}{\parbox{0.07\textwidth}{\sc {64-\\shot}}}} &
    
  Full FT + \textit{continued learning}  &  30.65 & 20.06 & 39.10 & 47.13 & 12.54 & 41.70 & 47.93 & 18.01 & 24.03 & 46.00 & ~~9.96
    \\
 & LoRA + \textit{continued learning}   &	\bf 42.07 & \bf 37.16 & \bf 53.76 & \bf 54.85 & 18.20 & 52.19 & \bf 52.90 & \bf 31.29 & \bf 37.10 & 52.61 & \bf 30.60
  \\
  & LoraHub
  & 
41.34 & 36.56 & 44.52 & 54.47 & \bf 47.70 & \bf 53.45 & 45.16 & 29.60 & 23.98 & \bf 54.35 & 23.65         \\

\bottomrule
\end{tabular}
}
\caption{Cross-lingual transfer results (NLI) on 10 XLSum
  languages for PaLM 2-XXS model. 16- and 64-shot experiments
  show average results from three different seed runs. For \textit{continued learning}, we use a 14/2 and 60/4 split for
  training/validation.}
 \label{tab:few-shot-xlsum-nli}
\end{table*} 

\begin{table*}[t]
\centering
\scalebox{0.8}{
\addtolength{\tabcolsep}{-2pt}
\begin{tabular}{ll|ccccccccccc}
\toprule
    
&\multicolumn{1}{l|}{\textbf{PaLM 2-XXS}}
& {\textbf{AVG}} & {{AZ}} & {{BN}} & {{JA}}& {{RN}} & {{KO}} & {{NE}} & {{GD}} & {{SO}} & {{TH}} & {{YO}} \\
 \midrule

{\multirow{3}{*}{\parbox{0.07\textwidth}{\sc {zero-\\shot}}}}

& Full FT & 13.71 & 12.38 & 12.97 & 24.94 & ~~9.43 & ~~8.40 & 30.71 & 13.57 & ~~5.18 & 12.57 & ~~6.99 \\
  & LoRA-4  & 24.21 & 25.40 & 36.15 & 48.70 & ~~4.15 & 36.86 & 32.84 & ~~7.10 & ~~9.02 & 36.71 & ~~5.17  \\
    & Avg. LoRA & \bf 32.44 & \bf 29.79 & \bf 41.96 & \bf 53.99 & \bf 20.28 & \bf 46.02 & \bf 39.28 & \bf 19.00 & \bf 17.15 & \bf 41.79 & 
    \bf 15.11
    \\
  \midrule
    
{\multirow{3}{*}{\parbox{0.07\textwidth}{\sc {16-\\shot}}}} &
    
 Full FT + \textit{continued learning} & 18.79 & 11.44 & 29.33 & 26.83 & ~~9.69 & 24.72 & 26.20 & ~~8.31 & 11.22 & 29.66 & 10.51
    \\       

  & LoRA-4 + \textit{continued learning}     & 
\bf 26.47 & 26.22 & \bf 37.43 & 37.68 & 14.53 & \bf 39.02 & \bf 32.29 & 13.57 & \bf 15.95 & 33.12 & \bf 14.91
  \\
  & LoraHub &
26.07 & \bf 26.35 & 33.69 & \bf 38.75 & \bf 17.56 & 34.90 & 29.20 & \bf 17.87 & 13.45 & \bf 38.99 & ~~9.92 \\
  
     \midrule
    
{\multirow{3}{*}{\parbox{0.07\textwidth}{\sc {64-\\shot}}}} &
    
  Full FT + \textit{continued learning}  & 19.57 & 13.88 & 26.64 & 29.87 & ~~7.77 & 27.07 & 28.51 & ~~8.78 & 11.80 & 32.63 & ~~8.77
    \\
 & LoRA-4 + \textit{continued learning}   & 27.66 & \bf 27.02 & \bf 37.05 & 40.43 & ~~9.67 & 34.20 & \bf 35.36 & 16.84 & \bf 18.37 & 38.63 & \bf 19.03
  \\
  & LoraHub
  & 
\bf 28.02 & 26.81 & 31.88 & \bf 46.24 & \bf 23.37 & \bf 37.88 & 33.13 & \bf 17.36 & 11.72 & \bf 39.58 & 12.17        \\

\bottomrule
\end{tabular}
}
\caption{Cross-lingual transfer results (\textsc{seahorse}) on 10 XLSum
  languages for PaLM 2-XXS model. 16- and 64-shot experiments
  show average results from three different seed runs. For \textit{continued learning}, we use a 14/2 and 60/4 split for
  training/validation.
  }
 \label{tab:few-shot-xlsum-sh}
\end{table*} 

\begin{table*}[t]
\centering
\scalebox{0.8}{
\addtolength{\tabcolsep}{0pt}
\begin{tabular}{ll|cccccc}
\toprule
    
&\multicolumn{1}{l|}{\textbf{PaLM 2-XXS}}    & \textbf{AVG}   & {CS}   & {DE}         & {EN}          & {FR}          & {ZH}          \\
 \midrule

\multicolumn{1}{l}{\multirow{3}{*}{\parbox{0.07\textwidth}{\sc {zero- shot}}}}
 & Full FT              & 20.22          & 17.16          & 25.31          & 23.47          & 22.60          & 12.56          \\
& LoRA-4              & \textbf{28.46} & \textbf{28.55} & \textbf{31.57} & \textbf{32.93} & \textbf{30.27} & 18.99          \\
& Avg. LoRA   & 26.93 & 24.68 & 27.45 & 32.19 & 30.68 & \bf 19.66 \\
                           \midrule
\multicolumn{1}{l}{\multirow{3}{*}{\parbox{0.07\textwidth}{\sc {16- shot}}}}       
& Full FT + \textit{continued learning}   & 26.90          & 22.53          & 29.23          & 30.50          & 26.16          & 26.11          \\
&  LoRA-4 + \textit{continued learning}    & \textbf{30.05} & \textbf{27.68} & \textbf{33.76} & 31.98 & \textbf{30.12} & \textbf{26.70} \\
& LoraHub   & 27.59 & 26.09 & 29.81 & \bf 32.70 & 29.10 & 20.25        \\
                               \midrule
\multicolumn{1}{l}{\multirow{3}{*}{\parbox{0.07\textwidth}{\sc {64- shot}}}}
& Full FT + \textit{continued learning}       & 28.73         & 26.45          & 30.17          & 32.24          & 28.86          & 25.95          \\
& LoRA-4 + \textit{continued learning}       & \textbf{31.08} & \textbf{28.97} & \textbf{34.09} & \textbf{33.11} & \textbf{30.99} & \textbf{28.24} \\
& LoraHub   & 27.66 & 26.05 & 29.82 & 33.00 & 29.20 & 20.25    
\\
\bottomrule
\end{tabular}
}
\caption{Cross-lingual transfer results (ROUGE-L) on XWikis using
  \textit{leave-one-out} training.
Few-shot results are averaged across three seed runs. 14/2 and 60/4 splits are used for training/validation in \textit{continued learning}.
}
 \label{tab:few-shot-xwikis}
\end{table*} 

\begin{table*}[t]
\centering
\scalebox{0.8}{
\addtolength{\tabcolsep}{0pt}
\begin{tabular}{ll|cccccc}
\toprule
    
&\multicolumn{1}{l|}{\textbf{PaLM 2-XXS}}    & \textbf{AVG}   & {CS}   & {DE}         & {EN}          & {FR}          & {ZH}          \\
 \midrule

\multicolumn{1}{l}{\multirow{3}{*}{\parbox{0.07\textwidth}{\sc {zero- shot}}}}
 & Full FT              & 30.17 & 33.30 & 28.60 & 34.73 & 24.07 & 30.14 \\
& LoRA-4              & 48.31 & \bf 52.80 & \bf 44.27 & 48.66 & 40.05 & 55.78 \\
& Avg. LoRA   & \bf 49.29 & 52.67 & 42.86 & \bf 50.34 & \bf 43.76 & \bf56.80 \\
                           \midrule
\multicolumn{1}{l}{\multirow{3}{*}{\parbox{0.07\textwidth}{\sc {16- shot}}}}       
& Full FT + \textit{continued learning}   & 34.17 & 26.93 & 31.51 & 43.51 & 26.73 & 42.17 \\
&  LoRA-4 + \textit{continued learning}    & 45.90 & 48.17 & 37.67 & \bf 49.65 & 39.30 & \bf 54.73 \\
& LoraHub   & \bf 47.45 & \bf 52.71 & \bf 41.32 & 48.09 & \bf 43.15 & 51.98 \\
                               \midrule
\multicolumn{1}{l}{\multirow{3}{*}{\parbox{0.07\textwidth}{\sc {64- shot}}}}
&Full FT + \textit{continued learning}       & 39.47 & 35.29 & 32.36 & 54.19 & 35.73 & 39.81 \\
& LoRA-4 + \textit{continued learning}       & 45.12 & 48.46 & 36.48 & \bf 50.75 & 38.31 & 51.62 \\
& LoraHub   & \bf 48.09 & \bf 52.74 & \bf 41.31 & 50.63 & \bf 42.72 & \bf 53.05    
\\
\bottomrule
\end{tabular}
}
\caption{Cross-lingual transfer results (NLI) on XWikis using
  \textit{leave-one-out} training.
Few-shot results are averaged across three seed runs. 14/2 and 60/4 splits are used for training/validation in \textit{continued learning.
}}
 \label{tab:few-shot-xwikis-nli}
\end{table*} 

\begin{table*}[t]
\centering
\scalebox{0.8}{
\addtolength{\tabcolsep}{0pt}
\begin{tabular}{ll|cccccc}
\toprule
    
&\multicolumn{1}{l|}{\textbf{PaLM 2-XXS}}    & \textbf{AVG}   & {CS}   & {DE}         & {EN}          & {FR}          & {ZH}          \\
 \midrule

\multicolumn{1}{l}{\multirow{3}{*}{\parbox{0.07\textwidth}{\sc {zero- shot}}}}
 & Full FT              &  16.26 & 13.13 & 19.74 & 18.81 & 16.90 & 12.73 \\
& LoRA-4               & 26.40 & 29.84 & \bf 28.59 & 27.22 & 27.23 & 19.11 \\
& Avg. LoRA   & \bf 26.86 & \bf 30.13 & 28.48 & \bf 28.07 & \bf 28.18 & \bf 19.44 \\
                           \midrule
\multicolumn{1}{l}{\multirow{3}{*}{\parbox{0.07\textwidth}{\sc {16- shot}}}}       
& Full FT + \textit{continued learning}   & 21.82 & 18.10 & 25.53 & 25.00 & 21.47 & 18.98 \\
&  LoRA-4 + \textit{continued learning}    & \bf 28.20 & 27.80 & \bf 29.59 &\bf  29.43 & \bf 29.18 & \bf 25.02 \\
& LoraHub   & 25.84 & \bf 28.95 & 27.48 & 28.20 & 27.63 & 16.96 \\
                               \midrule
\multicolumn{1}{l}{\multirow{3}{*}{\parbox{0.07\textwidth}{\sc {64- shot}}}}
& Full FT + \textit{continued learning}       & 24.16 & 22.92 & 23.31 & 29.15 & 26.80 & 18.64 \\
& LoRA-4 + \textit{continued learning}       & \bf 28.05 & 28.02 & \bf 28.80 & \bf 30.08 & \bf 29.16 & \bf 24.17 \\
& LoraHub   & 26.56 & \bf 29.34 & 27.33 & 29.93 & 27.60 & 18.58
\\
\bottomrule
\end{tabular}
}
\caption{
Cross-lingual transfer results (\textsc{seahorse}) on XWikis using
  \textit{leave-one-out} training.
Few-shot results are averaged across three seed runs. 14/2 and 60/4 splits are used for training/validation in \textit{continued learning}.
}
 \label{tab:few-shot-xwikis-sh}
\end{table*}

 \autoref{tab:xwikis-per-lang}, \autoref{tab:xwikis-per-lang-nli}, and \autoref{tab:xwikis-per-lang-sh}  show ROUGE-L, NLI, and \textsc{seahorse}
 results for PaLM 2-XXS on XWikis for individual
 languages;  in the high-data regime
 and in a zero-shot cross-lingual transfer setting from English.   \autoref{tab:xlsum-per-lang}, \autoref{tab:xlsum-per-lang-nli}, and \autoref{tab:xlsum-per-lang-sh}  show ROUGE-L, NLI, and \textsc{seahorse}
 results for PaLM 2-XXS on XLSum for individual
 languages;  in the high-data regime
 and in a zero-shot cross-lingual transfer setting from English.

 \begin{table}[t]
\centering

\addtolength{\tabcolsep}{0pt}
\scalebox{0.85}{
\begin{tabular}{l|cc|cc}
\toprule
\multirow{2}{*}{\textbf{PaLM 2-XXS}} & \multicolumn{2}{c|}{\textbf{High-Data}} & \multicolumn{2}{c}{\textbf{EN Zero-Shot}}                    \\
& FT & LoRA & FT & LoRA \\
\midrule
\bf Average & 34.08 & 32.92 & 17.51 & 23.86 \\ 
English     & 35.13 & 34.16 & --- & --- \\
German      & 36.97 & 36.08 & 20.64 & 27.89 \\
French      & 34.53 & 33.65 & 16.77 & 27.65 \\
Czech       & 31.92 & 30.82 & 19.21 & 26.11  \\
Chinese     & 31.82 & 29.91 & 13.43 & 13.77 \\
\bottomrule
\end{tabular}}
\caption{
Per language ROUGE-L results on XWikis  using full fine-tuning (FT) and
LoRA (rank 4), for PaLM 2-XXS  in the high-data regime and in a  zero-shot cross-lingual transfer
 setting from English.
}
\label{tab:xwikis-per-lang}
\end{table}

\begin{table}[t]
\centering

\addtolength{\tabcolsep}{0pt}
\scalebox{0.85}{
\begin{tabular}{l|cc|cc}
\toprule
\multirow{2}{*}{\textbf{PaLM 2-XXS}} & \multicolumn{2}{c|}{\textbf{High-Data}} & \multicolumn{2}{c}{\textbf{EN Zero-Shot}}                    \\
& FT & LoRA & FT & LoRA \\
\midrule
\bf Average & 41.04 & 47.43 & 35.95 & 45.54 \\ 
English     & 48.34 & 51.35 & --- & --- \\
German      & 35.27 & 39.42 & 37.09 & 42.02 \\
French      & 35.58 & 39.31 & 37.32 & 41.73 \\
Czech       & 42.46 & 50.78 & 33.75 & 51.75  \\
Chinese     & 43.53 & 56.30 & 35.66 & 46.65 \\
\bottomrule
\end{tabular}}
\caption{
Per language NLI results on XWikis  using full fine-tuning (FT) and
LoRA (rank 4), for PaLM 2-XXS  in the high-data regime and in a  zero-shot cross-lingual transfer
 setting from English.
}
\label{tab:xwikis-per-lang-nli}
\end{table}

\begin{table}[t]
\centering

\addtolength{\tabcolsep}{0pt}
\scalebox{0.85}{
\begin{tabular}{l|cc|cc}
\toprule
\multirow{2}{*}{\textbf{PaLM 2-XXS}} & \multicolumn{2}{c|}{\textbf{High-Data}} & \multicolumn{2}{c}{\textbf{EN Zero-Shot}}                    \\
& FT & LoRA & FT & LoRA \\
\midrule
\bf Average & 25.19 & 24.20 & 22.43 & 25.96 \\ 
English     & 27.80 & 29.37  & --- & --- \\
German      & 26.42 & 28.88 & 25.39 & 28.47 \\
French      & 25.95 & 29.15 & 24.87 & 28.46 \\
Czech       & 25.36 & 28.40 & 21.86 & 28.97  \\
Chinese     & 20.42 & 22.79 & 17.60 & 17.93 \\
\bottomrule
\end{tabular}}
\caption{
Per language \textsc{seahorse} results on XWikis  using full fine-tuning (FT) and
LoRA (rank 4), for PaLM 2-XXS  in the high-data regime and in a zero-shot cross-lingual transfer setting from English.
}
\label{tab:xwikis-per-lang-sh}
\end{table}

 \begin{table}[ht]
\centering

\addtolength{\tabcolsep}{-1.5pt}
\scalebox{0.85}{
\begin{tabular}{l|cc|cc}
\toprule
\multirow{2}{*}{\textbf{PaLM 2-XXS}} & \multicolumn{2}{c|}{\textbf{High-Data}} & \multicolumn{2}{c}{\textbf{EN Zero-Shot}}                    \\
& FT & LoRA & FT & LoRA \\
\midrule
\textbf{Average}    &  31.11  & 29.03& ~~5.20 &  21.13  \\ 
English             &  32.33 & 31.25 & \multicolumn{1}{c}{--}       & \multicolumn{1}{c}{~~~~--}   \\
Hindi               & 35.41 & 32.80 & ~~4.43 & 27.24 \\
Urdu                & 34.93 & 31.70 & ~~1.31 & 22.67  \\
Russian             & 27.40 & 25.08 & ~~5.67 & 22.60   \\
Portuguese          & 30.82 & 28.50 & ~~8.85 & 26.44  \\
Persian             & 34.64 & 31.91 & ~~3.67 & 27.94  \\
Ukrainian           & 27.64 & 24.92 & ~~5.82 & 19.32  \\
Indonesian          & 33.42 & 31.28 & ~~8.49 & 27.87  \\
Spanish             & 26.21 & 24.80 & ~~8.21 & 22.85   \\
Arabic              & 29.47 & 27.52 & ~~4.19 & 23.26  \\
Chinese-Traditional & 36.74 & 33.44 & ~~3.03  & 25.86 \\
Chinese-Simplified  & 36.95 & 33.96 & ~~2.11 & 28.54  \\
Vietnamese          & 32.11 & 30.00 & ~~5.68 & 24.69  \\
Turkish             & 31.09 & 28.08 & ~~7.11  & 24.41 \\
Tamil               & 32.71 & 29.67 & ~~3.44 & 21.22  \\
Pashto              & 36.41 & 33.81 & ~~3.13 & 14.76  \\
Marathi             & 28.14 & 26.25 & ~~4.73 & 17.97  \\
Telugu              & 29.62 & 27.31 & ~~4.04 & 19.30  \\
Welsh               & 30.72 & 27.72 & ~~6.78 & 23.75  \\
Pidgin              & 31.50 & 30.37 & 16.84 & 22.76  \\
Gujarati            & 35.79 & 33.43 & ~~3.88 & 26.00  \\
French              & 29.67 & 29.74 & 10.32 & 26.55  \\
Punjabi             & 42.01 & 40.61 & ~~2.08 & 34.14 \\
Bengali             & 29.52 & 28.26 & ~~1.85 & 22.29  \\
Swahili             & 31.11 & 29.81  & ~~6.45  & 25.14\\
Serbian-Latin       & 22.92 & 21.94 & ~~5.56 & 18.87  \\
Serbian-Cyrillic    & 24.55 & 23.60 & ~~5.49 & 15.28 \\
Japanese            & 38.34 & 36.08 & ~~2.01 & 29.04  \\
Thai                & 25.93 & 26.53 & ~~4.53 & 22.26   \\
Azerbaijani         & 24.01 & 23.36 & ~~5.34 & 14.31  \\
Hausa               & 33.03 & 28.85 & ~~7.55  & 15.82  \\
Yoruba              & 33.66 & 29.87 & ~~8.40 & 20.30  \\
Oromo               & 23.89 & 19.35 & ~~5.42 & ~~7.15 \\
Somali              & 26.31 & 24.56 & ~~6.33 & 18.14   \\
Nepali              & 32.28 & 30.81 & ~~1.79  & 23.07  \\
Amharic             & 36.45 & 34.61 & ~~1.76 & 11.22  \\
Kirundi             & 25.55 & 19.28 & ~~7.16 & ~~8.80  \\
Tigrinya            & 39.85 & 36.34 & ~~1.52 & 16.90 \\
Uzbek               & 24.21 & 23.09 & ~~3.51 & 12.39  \\
Burmese             & 35.79 & 33.33 & ~~1.63 & 23.66  \\
Korean              & 32.92 & 31.03 & ~~6.78 & 23.05   \\
Igbo                & 30.59 & 28.57 & ~~7.74 & 16.63  \\
Sinhala             & 35.75 & 35.26 & ~~2.93 & 27.88   \\
Kyrgyz              & 22.61 & 22.18 & ~~4.31 & 12.24   \\
Scottish-Gaelic     & 25.10 & 25.55 & ~~6.72 & 15.21 \\
\bottomrule
\end{tabular}}
\caption{Per language ROUGE-L results on XLSum  using full fine-tuning (FT) and
LoRA (rank 4), for PaLM 2-XXS  in the 
 high-data regime and in a zero-shot cross-lingual transfer
 setting from English.
}
\label{tab:xlsum-per-lang}
\end{table}

\begin{table}[ht]
\centering

\addtolength{\tabcolsep}{-1.5pt}
\scalebox{0.85}{
\begin{tabular}{l|cc|cc}
\toprule
\multirow{2}{*}{\textbf{PaLM 2-XXS}} & \multicolumn{2}{c|}{\textbf{High-Data}} & \multicolumn{2}{c}{\textbf{EN Zero-Shot}}                    \\
& FT & LoRA & FT & LoRA \\
\midrule
\textbf{Average}    & 42.93 & 51.16 & ~~4.49 &  39.07  \\
English             & 62.34 & 62.60 & \multicolumn{1}{c}{--}       & \multicolumn{1}{c}{~~~~--}   \\
Hindi               & 55.00 & 56.15 & ~~2.02 & 49.50 \\
Urdu                & 53.06 & 52.70 & ~~0.58 & 34.95  \\
Russian             & 50.39 & 51.32 & ~~4.63 & 51.37  \\
Portuguese          & 43.72 & 43.55 & 15.98 & 47.45  \\
Persian             & 61.11 & 62.91 & ~~1.71 & 61.88 \\
Ukrainian           & 47.34 & 50.00 & ~~3.35 & 34.03 \\
Indonesian          & 57.03 & 60.26 & ~~8.02 & 61.63  \\
Spanish             & 43.06 & 47.82 & 22.81 & 57.13  \\
Arabic              & 46.92 & 48.78 & ~~2.09 & 44.86  \\
Chinese-Traditional & 57.04 & 60.85 & ~~0.80 & 58.49 \\
Chinese-Simplified  & 57.13 & 59.94 & ~~1.84 & 57.74  \\
Vietnamese          & 55.83 & 59.04 & ~~3.40 & 54.95 \\
Turkish             & 46.84 & 51.02 & ~~5.44 & 44.23 \\
Tamil               & 61.37 & 63.26 & ~~1.59 & 42.65  \\
Pashto              & 45.49 & 46.87 & ~~1.97 & 24.39  \\
Marathi             & 45.66 & 55.50 & ~~3.12 & 42.45  \\
Telugu              & 43.73 & 50.49 & ~~0.88 & 33.70  \\
Welsh               & 41.63 & 46.40 & ~~2.33 & 35.87  \\
Pidgin              & 41.96 & 49.96 & 32.69 & 59.13  \\
Gujarati            & 44.91 & 52.15 & ~~0.46 & 32.48  \\
French              & 43.49 & 54.14 & 25.62 & 53.52 \\
Punjabi             & 33.92 & 45.35 & ~~0.49 & 26.78\\
Bengali             & 51.17 & 63.23 & ~~2.93 & 47.99  \\
Swahili             & 38.19 & 49.10 & ~~4.65 & 40.19\\
Serbian-Latin       & 35.07 & 44.41 & ~~4.91 & 44.72  \\
Serbian-Cyrillic    & 29.72 & 42.63 & ~~1.12 & 45.97 \\
Japanese            & 59.18 & 63.13 & ~~3.44 & 58.70  \\
Thai                & 42.09 & 53.77 & ~~1.42 & 52.30   \\
Azerbaijani         & 31.69 & 48.98 & ~~4.04 & 28.95 \\
Hausa               & 31.37 & 38.11 & ~~2.31 & 19.91  \\
Yoruba              & 34.71 & 42.99 & ~~6.02 & 18.23  \\
Oromo               & 37.75 & 57.54 & ~~3.70 & 17.95\\
Somali              & 31.43 & 42.26 & ~~4.69 & 22.16   \\
Nepali              & 46.71 & 59.99 & ~~0.54 & 40.40 \\
Amharic             & 35.94 & 52.42 & ~~0.18 & 28.72  \\
Kirundi             & 22.97 & 28.42 & ~~3.45 & 18.49 \\
Tigrinya            & 41.04 & 44.01 &~~0.53 & 28.01 \\
Uzbek               & 29.78 & 46.71 &~~2.14 & 17.78  \\
Burmese             & 36.21 & 54.03 & ~~1.00 & 32.02  \\
Korean              & 47.31 & 62.30 & ~~2.72 & 48.97   \\
Igbo                & 32.43 & 38.73 & ~~3.76 & 23.81  \\
Sinhala             & 31.49 & 57.47 & ~~0.17 & 34.10   \\
Kyrgyz              & 27.39 & 47.95 & ~~0.70 & 26.43   \\
Scottish-Gaelic     & 19.42 & 32.90 & ~~1.46 & 13.94 \\

\bottomrule
\end{tabular}}
\caption{Per language NLI results on XLSum  using full fine-tuning (FT) and
LoRA (rank 4), for PaLM 2-XXS  in the 
 high-data regime and in a  zero-shot cross-lingual transfer
 setting from English.
}
\label{tab:xlsum-per-lang-nli}
\end{table}

\begin{table}[ht]
\centering

\addtolength{\tabcolsep}{-1.5pt}
\scalebox{0.85}{
\begin{tabular}{l|cc|cc}
\toprule
\multirow{2}{*}{\textbf{PaLM 2-XXS}} & \multicolumn{2}{c|}{\textbf{High-Data}} & \multicolumn{2}{c}{\textbf{EN Zero-Shot}}                    \\
& FT & LoRA & FT & LoRA \\
\midrule
\textbf{Average}    & 31.64 & 34.42 & 6.88 & 23.08  \\
English             &  42.00 & 42.72 & \multicolumn{1}{c}{--}       & \multicolumn{1}{c}{~~~~--}   \\
Hindi               & 39.78 & 40.39 & ~~5.51 & 34.22 \\
Urdu                & 40.96 & 38.77 & ~~4.28 & 20.58  \\
Russian             & 43.60 & 44.32 & ~~6.79 & 41.20  \\
Portuguese          & 33.19 & 34.56 & 12.95 & 37.87  \\
Persian             & 44.39 & 45.43 & ~~5.68 & 42.24 \\
Ukrainian           & 40.90 & 43.22 & ~~6.29 & 26.24 \\
Indonesian          & 41.22 & 44.43 & ~~7.04 & 40.13  \\
Spanish             & 33.30 & 37.06 & 19.19 & 45.30  \\
Arabic              & 36.35 & 38.57 & ~~5.09 & 32.96  \\
Chinese-Traditional & 41.70 & 43.34 & ~~5.09 & 35.71 \\
Chinese-Simplified  & 41.68 & 42.84 & ~~5.10 & 39.12  \\
Vietnamese          & 35.78 & 36.42 & ~~6.03 & 28.99 \\
Turkish             & 46.85 & 48.85 & ~~7.19 & 44.01 \\
Tamil               & 38.00 & 37.06 & ~~4.43 & 21.96  \\
Pashto              & 29.78 & 25.35 & ~~4.10 & ~~5.60  \\
Marathi             & 33.08 & 35.48 & ~~7.82 & 25.43  \\
Telugu              & 25.97 & 26.26 & ~~5.49 & 10.30  \\
Welsh               & 29.00 & 28.98 & ~~5.93 & 17.44 \\
Pidgin              & 28.82 & 32.71 & 22.28 & 37.42  \\
Gujarati            & 26.13 & 27.99 & ~~5.56 & 13.49 \\
French              & 36.77 & 48.56 & 20.38 & 43.65 \\
Punjabi             & 22.58 & 23.84 & ~~4.80 & 10.53 \\
Bengali             & 38.57 & 43.09 & ~~7.72 & 32.00  \\
Swahili             & 31.57 & 39.15 & ~~5.72 & 28.45 \\
Serbian-Latin       & 29.87 & 38.37 & ~~9.78 & 36.45 \\
Serbian-Cyrillic    & 24.71 & 32.70 & ~~4.92 & 22.66 \\
Japanese            & 38.68 & 43.81 & ~~7.77 & 45.05  \\
Thai                & 31.35 & 38.99 & ~~5.68 & 32.21   \\
Azerbaijani         & 27.07 & 35.27 & ~~5.29 & 15.41 \\
Hausa               & 25.95 & 28.11 & ~~4.71 & ~~8.14  \\
Yoruba              & 26.45 & 26.82 & ~~5.50 & ~~7.52 \\
Oromo               & 24.63 & 22.66 & ~~6.16 & ~~6.26 \\
Somali              & 23.87 & 22.36 & ~~5.01 & ~~7.59   \\
Nepali              & 33.83 & 40.11 & ~~5.11 & 23.67 \\
Amharic             & 25.44 & 26.81 & ~~4.79 & ~~4.67  \\
Kirundi             & 18.69 & 14.30 & ~~5.89 & ~~6.64 \\
Tigrinya            & 26.61 & 17.19 & ~~4.71 & ~~5.19 \\
Uzbek               & 25.59 & 30.67 & ~~5.17 & ~~6.05  \\
Burmese             & 28.06 & 35.06 & ~~4.92 & 10.54  \\
Korean              & 32.52 & 40.15 & ~~5.95 & 27.97   \\
Igbo                & 18.25 & 19.65 & ~~5.92 & ~~7.27  \\
Sinhala             & 24.01 & 34.99 & ~~4.62 & 13.99   \\
Kyrgyz              & 20.81 & 29.62 & ~~5.22 & ~~8.29  \\
Scottish-Gaelic     & 15.65 & 22.03 & ~~4.95 & ~~5.05 \\

\bottomrule
\end{tabular}}
\caption{Per language \textsc{seahorse} results on XLSum  using full fine-tuning (FT) and
LoRA (rank 4), for PaLM 2-XXS  in the 
 high-data regime and in a  zero-shot cross-lingual transfer
 setting from English.
}
\label{tab:xlsum-per-lang-sh}
\end{table}
 
\section{Examples of Summaries}
 \label{examples}
We provide some randomly selected examples of input, target, and generated summaries of XLSum with Full FT and LoRA-4 models in \autoref{tab:examples}.

\clearpage

\begin{table*}[!ht]
\centering
\
\scalebox{0.8}{
\addtolength{\tabcolsep}{0pt}
\begin{tabular} {p{2cm}|p{16cm}}
\toprule
\textsc{Language} & \textsc{English}   \\
\midrule
\sc Input & Sir Charlie Mayfield said he "didn't say Brexit was the reason" for a 99\% slide in half-year profits. "The fact is sterling is weaker, it's more expensive to import goods... so we have to absorb that within our margin," he said. "I'm not going to get into some sort of ding-dong with the secretary of state." Dominic Raab told the BBC on Thursday it was a mistake for "business that aren't doing so well to blame Brexit". "I don't doubt that some of the uncertainty around these negotiations will have an impact on business - that's why we are putting all our energy into getting the good deal we want with our EU friends and partners," he said. "All I am just gently saying is that it's rather easy for a business to blame Brexit and the politicians rather than take responsibility for their own situation." Mr Raab's comments appear to have been in response to Sir Charlie's warning in John Lewis's half-year results that "with the level of uncertainty facing consumers and the economy, in part due to ongoing Brexit negotiations, forecasting is particularly difficult". The owner of the John Lewis department store chain and Waitrose said it continued to expect annual profits to be "substantially lower than last year". Mr Raab also said the government was preparing for a no-deal Brexit despite being confident that eventuality would not come to pass: "Getting a deal with the European Union is still by far and away the most likely outcome." This week Ralf Speth, the boss of Jaguar Land Rover, warned the government needed to get "the right Brexit" or risk wiping out profits at the UK's biggest carmaker and trigger big job cuts. Meanwhile, the Brexit secretary welcomed a promise by two mobile operators, Vodafone and Three, not to impose Europe roaming charges for UK customers if Britain leaves the bloc with no deal. "What we have said is we would like to see other companies following suit, but, in any event, we would legislate for a limit on roaming charges to make sure in a no-deal scenario that we protect British consumers," Mr Raab said. A new raft of technical papers is being released by the government on Thursday outlining the impact of a no-deal Brexit on business and consumers. Mr Raab also accused those warning about shortages of food and medicines after a no-deal withdrawal of "scaremongering", saying it was "nonsense" to claim UK supermarkets would run out of food.\\
\midrule
\sc Target & The boss of John Lewis has hit out at Brexit secretary Dominic Raab after he said companies should not blame Brexit for poor results. \\
\midrule
\sc Full FT &  The boss of John Lewis has defended his company's decision to blame Brexit for its falling profits.\\
\midrule
\sc LoRA-4 &  The boss of John Lewis has said it is "rather easy" for businesses to blame Brexit for their poor performance.
\\
\midrule
\midrule
\sc Input &On Tuesday, the corporation's director general Tony Hall will announce new commissions including a series in collaboration with the Tate. Lord Hall will say the arts are at risk of becoming marginalised for future generations unless more is done to get children and young people engaged. As part of the plans, acclaimed 1960s arts show Civilisation is to return. The landmark 13-part series series telling the history of western art was fronted by art historian Kenneth Clark when it was screened by BBC Two in 1969. The BBC is now looking for a new authoritative presenter to follow in Clark's footsteps for the new series, which will use modern digital technology to provide a fresh overview of the history of art. The BBC said: "Civilisation was a landmark programme. Producing something of equal scale will be a huge challenge, but we believe it is time to once again create something with the same impact for a new generation. "The arts have been central to the BBC's past and are central to its future. As all arts organisations face the challenge of delivering more in a tight economic climate, it is vital that we work together in new ways to create a bigger and better offer to the public." As well as including arts content in The One Show, the idea will be spread to other popular programmes such as BBC Radio 2's Simon Mayo show. Lord Hall announced last October that funding for the BBC's arts coverage would increase by 20\%. The director general, who was Royal Opera House chief executive for 12 years before his appointment to the BBC, said he wanted to return the arts to the corporation's "heart". At that time, he announced that Simon Schama would front a five-part series in partnership with the National Portrait Gallery exploring the history of Britain through portraiture, and that Andrew Marr would present a new series looking at the greatest writers in Scotland.\\
\midrule
\sc Target &  The BBC is to feature more arts stories in programmes like The One Show in a bid to make the subject less elitist. \\
\midrule
\sc Full FT &  The BBC is to include arts content in its The One Show, as part of a new £100m arts programme.\\
\midrule
\sc LoRA-4 & The BBC is to launch a new arts programme in The One Show, which will be fronted by a new presenter.
\\
\bottomrule
\end{tabular}
}
\end{table*}

\clearpage

\begin{table*}[!ht]
\centering
\scalebox{0.8}{
\addtolength{\tabcolsep}{0pt}
\begin{tabular} {p{2cm}|p{16cm}}
\toprule
\textsc{Language} & \textsc{Simplified Chinese}   \\
\midrule
\sc Input & \begin{CJK}{UTF8}{gbsn}这对新人周五（7月17日）在温莎的诸圣皇家礼拜堂（The Royal Chapel of All Saints）举行了小型仪式，英国王室隔天公布了两张官方照片。 其中一张可见二人笑容满面，在鲜花搭制的拱门下离开教堂。另一张则是一对新人在教堂外与碧翠丝的祖父母——女王及爱丁堡公爵菲利普亲王（Prince Philip, Duke of Edinburgh）合照。 碧翠丝的父母约克公爵及公爵夫人（Duke and Duchess of York）并没有出现在官方公开的照片中，但白金汉宫证实，约克公爵安德鲁王子（Prince Andrew）牵着碧翠丝的手走进教堂。安德鲁王子在他的前好友麦克斯维尔（Ghislaine Maxwell）最近因涉嫌性贩卖遭逮捕后一直保持低调，麦克斯维尔对自己遭到的指控予以否认。 为了此次婚礼，女王将一条复古长裙和她在1947年自己婚礼上佩戴的一顶穗状钻石王冠借给了碧翠丝。 碧翠丝与莫奇原计划于今年5月举行婚礼，但受COVID-19新型冠状病毒疫情影响，二人决定将仪式延后，并改为举行由双方父母及兄弟姐妹参加的私人典礼。 英国自3月23日起实行全国范围的封锁，英格兰境内几乎所有婚礼均被禁止举行。而自7月4日起，英格兰开始允许举行婚礼，但人数上限为30人。 白金汉宫在一则声明中表示，碧翠丝的婚礼符合政府政策规定。 这应该是94岁的女王与99岁的菲利普亲王在英国封锁令后首次参加家庭聚会。 碧翠丝在婚礼上身穿一条已故设计师哈特内尔（Norman Hartnell）设计的象牙色双面横绫缎塔夫绸长裙，配以公爵夫人缎并镶有钻石，这条长裙借自女王。 碧翠丝婚纱所用长裙是女王在1962年12月出席电影《沙漠枭雄》（Lawrence Of Arabia）全球首映式时所穿的礼服的改良版。 白金汉宫称，这条长裙经过了女王的高级服装师安吉拉·凯利（Angela Kelly）及设计师斯图尔特·帕文（Stewart Parvin）改良。 碧翠丝是目前英国王位第九顺位继承人，她在婚礼上佩戴的玛丽女王（Queen Mary）穗状钻石王冠也借自她的祖母女王。 1947年11月，女王（时为伊丽莎白公主）与菲利普王子结婚，当时所佩戴的王冠与这次碧翠丝婚礼所用的是同一顶王冠。 今年31岁的碧翠丝与35岁的莫奇原计划于今年5月29日在伦敦的圣詹姆斯宫（St James Palace）皇家礼拜堂（Chapel Royal）举行婚礼。最终仪式场所所在的温莎大公园（Windsor Great Park）皇家小屋（Royal Lodge）是碧翠丝童年时代的家，她的父母如今仍然居住在那里。 碧翠丝与她的丈夫自2018秋天起交往，去年9月订婚。据称这对新人是在碧翠丝的妹妹尤金妮公主（Princess Eugenie）的婚礼上重逢后开始相恋。 婚礼结束后，新娘一般会举行“抛花球”仪式。但遵循王室新娘传统，碧翠丝的捧花被放置在西敏寺大教堂（Westminster Abbey）的无名战士墓碑上。 这束手捧花由茉莉、淡粉及奶油色的香豌豆花、王室陶瓷象牙白多头玫瑰、粉色庭园玫瑰、粉蜡花、浅粉色落新妇花组合而成，并按照王室传统加入多支紫薇花。 西敏寺大教堂法政牧师安东尼·波尔（The Reverend Canon Anthony Ball）及皇家礼拜堂副主任牧师保罗·怀特（Reverend Paul Wright）的儿子托比·怀特（Toby Wright）一同将捧花放置在威斯敏斯特教堂中。 文中图片，版权所有，未经批准，翻印必究。\end{CJK}\\
\midrule
\sc Target &  \begin{CJK}{UTF8}{gbsn}英国碧翠丝公主（Princess Beatrice）与地产大亨爱德华多·马佩利·莫奇（Edoardo Mapelli Mozzi）举行婚礼。白金汉宫公开了婚礼现场照片。\end{CJK} \\
\midrule
\sc Full FT &  \begin{CJK}{UTF8}{gbsn}英国女王伊丽莎白二世（Queen Elizabeth II）的孙女碧翠丝（Bessie Louise Mountbatten-Windsor）与美国商人莫奇·艾米斯（Mitch Eisley）在温莎举行了一场私人仪式，女王将借此机会借给碧翠丝一顶穗状钻石王冠。\end{CJK}\\
\midrule
\sc LoRA-4 &  \begin{CJK}{UTF8}{gbsn}英国王室宣布，英国王位第九顺位继承人碧翠丝·莫奇（Princess Beatrice）与丈夫莫奇·莫奇（Mohamed bint Mohammed bin Rashid Al Maktoum）在温莎的皇家小屋举行了私人婚礼。\end{CJK}
\\
\midrule
\midrule
\sc Input & \begin{CJK}{UTF8}{gbsn}考纳利在伦敦奥运开幕式上做一把女王替身。 来自泰晤士河边亨利地区的考纳利(Gary Connery)，在伦敦奥运会开幕式上，身着女王的服饰从直升机跳下。他凭借此举，与12位候选人共同入围一项电视年度探险大奖。 另外，从太空跳伞成功的奥地利人波姆加德纳，同样栖身候选名单。 今年五月，在没有降落伞的情况下，考纳利从直升机上跳下，并成功落入一堆空纸盒中。 这位43岁的冒险家在说到做女王替身、和扮演邦德的克雷格一起完成这个伦敦奥运开幕式的惊心动魄的节目时表示，这是个“极其精彩的生活经历”。 他说：“当我回顾今年时，我会在脸上露出绽放的笑容。” 考纳利说，让他颇为感动的是，很多英国民众对他的冒险行为都非常支持。他说：“民众的反应让我欣慰。” 他说，不管最终能不能获奖，能与这些一年来活得精彩的人一起入围，“就是份荣耀”。 与内文相关的链接 相关话题。\end{CJK}\\
\midrule
\sc Target &  \begin{CJK}{UTF8}{gbsn}在伦敦奥运会开幕式表演中做女王替身的考纳利，已经获得一项探险大奖的提名。\end{CJK} \\
\midrule
\sc Full FT &  \begin{CJK}{UTF8}{gbsn}英国一名男子冒险家，在伦敦奥运会开幕式上做一把女王替身，成功入围电视探险大奖。\end{CJK}\\
\midrule
\sc LoRA-4 &  \begin{CJK}{UTF8}{gbsn}英国一名冒险家在伦敦奥运开幕式上做一把女王替身，成为今年电视年度探险大奖的候选人。\end{CJK}
\\
\bottomrule
\end{tabular}
}
\end{table*}

\clearpage

\begin{table*}[!ht]
\centering
\scalebox{0.8}{
\addtolength{\tabcolsep}{0pt}
\begin{tabular} {p{2cm}|p{16cm}}
\toprule
\textsc{Language} & \textsc{French}   \\
\midrule
\sc Input & Les logiciels de rançonnage ont rapporté plus 25 millions de dollars lors des deux précédentes années, selon Google. Le géant de la recherche sur internet a créé des milliers de victimes virtuelles des logiciels de rançonnage pour les besoins d'une étude. Son but était d'exposer le processus de paiement dans l'utilisation de ces programmes utilisés par les cybercriminels. La plupart d'entre eux ont gagné le plus d'argent en 2016. Il s'y sont mis après s'être rendu compte à quel point les logiciels de rançonnage pouvaient être rentables. Lire aussi: Qu'est-ce qu'est un logiciel de rançonnage Les paiements se font souvent en Bitcoin, une monnaie virtuelle Un logiciel de rançonnage est un programme malveillant qui affecte un ordinateur puis il crypte les fichiers de manière à ce qu'ils ne puissent plus être utilisés ou lus. Les données ne sont décryptées qu'une fois que la victime paie une rançon. Les paiements se font souvent en Bitcoin, une monnaie virtuelle. Suivre aussi: Younoussa Sanfo, expert en sécurité informatique, est notre invité.\\
\midrule
\sc Target &Les logiciels de rançonnage ont rapporté plus 25 millions de dollars lors des deux précédentes années, selon Google. \\
\midrule
\sc Full FT & Les logiciels de rançonnage ont rapporté plus 25 millions de dollars lors des deux dernières années, selon Google.\\
\midrule
\sc LoRA-4 &  Google a publié une étude sur les logiciels de rançonnage.
\\
\midrule
\midrule
\sc Input &Les membres de la Cour suprême du Kenya qui ont statué vendredi sur l'annulation de la présidentielle du 8 août dernier L'Association des juges et magistrats du Kenya (KMJA) "s'insurge contre les remarques" de M. Kenyatta, "qui ont eu pour effet de dénigrer les juges de la Cour Suprême", a affirmé samedi soir Bryan Khaemba, secrétaire général de cette association. Après avoir dans un premier temps adopté un ton conciliant face au verdict rendu vendredi par la Cour suprême, le chef de l'Etat a rapidement durci le ton. Lors d'un rassemblement impromptu vendredi après-midi avec des partisans à Nairobi, il a d'abord qualifié les juges de la Cour suprême d'"escrocs". Puis s'adressant samedi à des représentants de son parti, il a signalé que cela ne peut plus durer. Lire aussi: "Nous nous pencherons sur ce problème, après les élections. Il y a un problème et nous devons le régler", a-t-il mis en garde. Un nouveau scrutin présidentiel est prévu d'ici au 31 octobre.\\
\midrule
\sc Target &  Au moins 73 morts et 110 blessés ont été dénombrés dans l'explosion d'un camion-citerne, dans l'ouest du Mozambique, selon les autorités de ce pays. \\
\midrule
\sc Full FT &  L'explosion à l'origine de la mort de 73 personnes a eu lieu dans la province de Tete, dans le sud-est du Mozambique.\\
\midrule
\sc LoRA-4 & Les autorités mozambicaines ont annoncé vendredi que 73 personnes avaient été tuées dans une explosion qui a fait 100 blessés.
\\
\midrule
\midrule
\sc Input &Premier League : West Ham battu par Tottenham 2-3 A l'heure de jeu, le match semblait être plié quand Eriksen marque le troisième but de Tottenham. West Ham réduit le score à la 65ème minute par Chicharito. Aurier, l'arrière du PSG qui est passé à Tottenham cette saison, sera exclu après un deuxième carton jaune. West Ham commence à y croire, surtout à l'issue du second but marqué par le Sénégalais Kouyaté, mais le score en restera là.\\
\midrule
\sc Target &  Les deux premiers buts de Tottenham ont été marqués par Harry Kane (35e, 38e). \\
\midrule
\sc Full FT &  Tottenham s'est imposé devant West Ham 2-3.\\
\midrule
\sc LoRA-4 &Tottenham a battu West Ham 2-3 dans le match de la 17ème journée de la Premier League.
\\
\bottomrule
\end{tabular}
}
\end{table*}
\clearpage

\begin{table*}[!ht]
\centering
\scalebox{0.8}{
\addtolength{\tabcolsep}{0pt}
\begin{tabular} {p{2cm}|p{16cm}}
\toprule
\textsc{Language} & \textsc{Hausa}   \\
\midrule
\sc Input & Rabon da Bale ya yi wa Madrid wasa tun 5 ga Oktoba A tsakiyar mako ne dai aka ga Bale dan kasar Wales da sauran abokan wasansa a tsakiyar fili bayan wasa da ya buga wa kasar ta dauke da tutar an rubuta 'Wales. Golf. Madrid. In that order'. Rubutun yana lissafa abubuwan da Bale ya fi kauna ne wato: ya fi son buga wa kasarsa Wales wasa sannan wasan Golf kafin Real Madrid. (Wales, Golf. Madrid. A jere) Magoya bayan Real Madrid sun yi wa sunan dan wasan gaban ihu bayan an sanar da sunayen 'yan tawagar da za su buga wasan ana gab da take wasa. Sun sake yi masa ihun a lokacin da aka taso shi daga benci a minti na 67 - wasansa na farko kenan tun ranar 5 ga watan Oktoba. Bale ya fuskanci kakkausar suka daga kafafen yada labarai a kasar Spain bisa abin da ya aikata yayin da ya taimaka wa Wales ta samu gurbin shiga gasar Euro 2020 ta kasashen nahiyar Turai. Karim Benzema ne ya farke kwallon da Willian Jose ya jefa a ragar Madrid a minti na 37 kafin Valverde ya kara ta biyu, sai kuma Modric da ya kara ta ukun. Tun farko dai Barcelona ce ta fara cin wasanta a gidan Leganes. Yanzu Barcelona da Real Madrid ne ke saman teburin La Ligar da maki 28 kowaccensu, sai Atletico Madrid da take biye masu da maki 25. Koci Zinedin Zidane ya dage kan cewa ba sai an ci gaba da tattauna lamarin ba kuma bayan tashi daga wasan Bale ya samu goyon baya. An rika tafa masa bisa taimakon da ya bai wa Luka Modric yayin cin kwallo ta uku da kuma kokarinsa na tsare gida. Yanzu hankali ya karkata kan gasar Champions League yayin da Madrid din za ta karbi bakuncin Paris St-Germain ranar Talata da karfe 8:00 na dare.
\\
\midrule
\sc Target &  Magoya bayan Real Madrid na ta cecekuce kan Gareth Bale bayan ya shigo wasan da Real din ta lallasa Real Sociedad da 3-1 a wasan La Liga mako na 14 a filin wasa na Bernabéu.
\\
\midrule
\sc Full FT &  Gareth Bale ya zama dan wasa na biyu da ya buga wa Real Madrid wasa cikin mako uku, bayan da ya taimaka wa Wales ta samu gurbin shiga gasar Champions League.
\\
\midrule
\sc LoRA-4 &Bale ya shaida wa BBC cewa ya fi son buga wasan Golf kafin wasan Real Madrid da ya yi a ranar Lahadi.
\\
\midrule
\midrule
\sc Input &  Mawaki King Monada wanda ke sanye da hula ne ya rera wakar Ranar Talatar da ta gabata ne wani bidiyo ya fito da ke nuna wasu 'yan mata sanye da hijabi na tikar rawar da ta samo asali daga kasar Afirka ta Kudu. Wasu mutane sun ce 'yan matan da ke cikin bidiyon 'yan jihar Kano ne, amma kakakin hukumar Hisbah a Kano, Adamu Yahaya ya shaida wa BBC cewa ba su samu wani rahoto mai kama da wannan ba. "Gaskiyar maganar ita ce ba mu da masaniya game da wannan sabuwar rawar da ka ke magana a kai, amma aikinmu ne hana aikata laifuka masu bata al'adu da tarbiyyar al'umma. "Saboda haka idan muka sami rahoton da ke cewa wasu na yin irin wannan rawar, to ko shakka babu za mu kama su". Wannan sabuwar rawar ta samo asali ne daga wata waka da mawakin Afirka ta Kudu, King Monada, ya rera mai suna 'Maldwedhe' da aka ce tana caza kwakwalwar matasa. Ma'anar kalmar Maldwedhe ita ce cuta, kuma Monada ya yi amfani da kalmar ce dangane da wani maras lafiya da ke dauke da cutar wadda ke sa shi ya suma idan ya kama masoyiyarsa tare da wani. Ga dai yadda rawar ta Maldwedhe take: Ko a farkon shekarar 2017 ma an fito da sarar wata sabuwar rawa da matasa suka dinga yayinta a Najeriya, wato rawar dab, inda suke yi tamkar masu ruku'u su kuma rufe fuskarsu da hannu daya. A wancan lokacin ma an yi ta ce-ce-ku-ce a kan wannan dabi'a. Karanta karin wasu labarai masu alaka.
\\
\midrule
\sc Target &  Hukumar Hisbah ta jihar Kano ta ce za ta kama duk wanda aka samu yana taka sabuwar rawar 'fadi ka mutu' da ake yayinta a yanzu.
\\
\midrule
\sc Full FT &  Hukumar Hisbah a jihar Kano da ke arewacin Najeriya ta ce ba za ta kama 'yan mata da ke yin wata sabuwar rawa a jihar ba, wanda ake kira 'hannu babu tsohon'.
\\
\midrule
\sc LoRA-4 &Hukumar Hisbah ta jihar Kano ta ce ba su da wani rahoto game da wani bidiyo da aka tsara da ke nuna 'yan mata sanye da hijabi na tikar rawar da ta samo asali daga kasar Afirka ta Kudu. 
\\
\bottomrule
\end{tabular}
}
\end{table*}
\clearpage

\begin{table*}[!ht]
\centering
\scalebox{0.8}{
\addtolength{\tabcolsep}{0pt}
\begin{tabular} {p{2cm}|p{16cm}}
\toprule
\textsc{Language} & \textsc{Indonesian}   \\
\midrule
\sc Input & Andrew Marr, penyiar BBC yang berusia 50an dan rajin berolahraga lari, tidak memenuhi kategori itu. Usia adalah salah satu faktor risiko terbesar, tapi siapa saja dari usia berapa saja dapat terserang stroke. Lebih dari 150.000 orang di Inggris terserang stroke setiap tahun dan seperempat dari mereka berusia di bawah 65 tahun. Bahkan ada pula kasus stroke pada usia kanak-kanak. Banyak pilihan gaya hidup berisiko meningkatkan peluang terkena stroke. Merokok, kelebihan lemak di perut dan terlalu banyak mengonsumsi alkohol tidak serta merta mengakibatkan stroke, namun berangsur-angsur meningkatkan risiko seumur hidup. Namun ada penyebab lain stroke pada orang berusia muda dan sehat. Kelainan saat lahir Stroke menyebabkan otak kekurangan oksigen ketika aliran darah terhenti, baik karena penyumbatan (stroke ischaemic) atau oleh ledakan pembuluh darah di otak (stroke haemmorrhagic). Sekitar 80\% stroke disebabkan oleh penyumbatan darah, tapi stroke jenis apa yang menimpa Andrew Marr belum diketahui.l Namun jika anda melihat orang-orang dibawah usia 65, maka stroke haemorrhagic menjadi lebih umum. Stroke tipe ini dapat diakibatkan kelainan dalam pembuluh darah yang sudah ada sejak lahir. Bom waktu di otak ini dapat meledak sewaktu-waktu. Salah satu contoh adalah kelainan bentuk anarteriovenous, ketika arteri tersumbat yang berarti tekanan di dalam pembuluh darah terlalu besar untuk diatasi oleh tubuh. Akibatnya otak mengalami pendarahan. Bahkan stress dapat menaikkan tekanan darah cukup untuk memicu stroke dan ada beragam bukti mengenai dampak konsumsi kopi dalam jumlah terlalu banyak. Detak jantung yang tidak teratur atau atrial fibrillation juga mengakibatkan stroke ischaemic. Sebagian jantung berdetak begitu cepat sehingga organ ini berhenti bekerja sebagai pompa. Darah berkumpul di dalam jantung, yang dapat tersumbat, mengalir ke otak dan menyebabkan stroke. Satu faktor yang tidak dapat dihindari adalah gen. Ada orang yang peluang terserang stroke lebih besar dibandingkan yang lain dan hal itu biasanya diturunkan dalam satu keluarga. Salah kaprah Dr Clare Walton dari Asosiasi Stroke, mengatakan, "Saya akan mengatakan bahwa ada salah kaprah bahwa ini hanyalah kondisi orang tua. Seperempat kasus stroke terjadi pada orang usia kerja dan anak-anak serta bayi juga mengalami stroke. "Kita harus paham bahwa semua orang memiliki risiko terserang stroke dan bukan hanya di usia tua." Pada akhirnya semua tergantung pada sikap kita. Diet sehat, olahraga teratur, minum alkohol secukupnya saja dan tidak merokok secara dramatis mengurangi risiko stroke. Namun beberapa orang dengan gaya hidup sehat akan tetap terkena stroke, sedangkan mereka dengan gaya hidup sebaliknya justru tidak.
\\
\midrule
\sc Target & Saat anda berpikir mengenai pasien stroke "tipikal", anda langsung teringat pada seseorang yang berusia lanjut, kelebihan berat badan, malas berolahraga dan merokok. 
\\
\midrule
\sc Full FT &  Anda berusia 65 tahun dan Anda tidak memiliki risiko tinggi terkena stroke?
\\
\midrule
\sc LoRA-4 & Anda berusia 30 tahun, Anda sehat dan Anda tidak memiliki penyakit apa-apa. Anda mungkin tidak termasuk dalam kategori orang yang paling berisiko terkena stroke.
\\
\midrule
\midrule
\sc Input & Dua anak-anak Belanda diyakini berada di Raqqa wilayah kekuasaan ISIS Perempuan, yang telah bercerai dengan suaminya, itu membawa seorang anak laki-laki yang berusia delapan tahun dan anak perempuan berusia tujuh tahun diyakini telah bepergian menggunakan paspor palsu. Mantan suami yang merupakan ayah anak-anak, seorang warga negara Belanda, telah memperingatkan otoritas tentang kepergian mereka ke wilayah yang dikuasai ISIS. Otoritas mengatakan kasus ini merupakan yang pertama kali terjadi. Perempuan yang berusia 32 tahun itu, yang tidak disebutkan identitasnya, telah tinggal di bagian selatan kota Maastricht, Belanda. Dia dan kedua anaknya belum terlihat sejak Oktober lalu. Mereka diyakini terbang dengan pesawat ke Belgia ke Athena dan perempuan itu dilaporkan menghubungi ibunya pada Januari, dengan mengatakan mereka berada di wilayah kekuasaan ISIS di Raqqa di bagian utara Suriah. Jaksa memperlakukan kasus ini sebagai sebuah penculikan dan telah menerbitkan surat penahanan internasional. Tetapi mereka mengakui kita perempuan dan anak-anaknya telah menyebrang perbatasan ke Suriah, sangat kecil kemungkinan untuk membawa keluarga tersebut kembali ke Belanda. Sekitar 200 orang warga Belanda, termasuk sejumlah anak-anak, diketahui telah bergabung dengan ISIS di Irak dan Suriah. ISIS menjadi salah satu dari kelompok jihadis yang berbahaya, dituduh melakukan pembunuhan massal dan penghukum etnis dan pemeluk agama minoritas di wilayah yang mereka kuasai.
\\
\midrule
\sc Target &  Seorang perempuan Checnya yang tinggal di Belanda telah membawa dua anaknya untuk bergabung dengan militan Negara Islam ISIS di Suriah, yang bertentangan dengan keinginan ayah kedua anak tersebut, seperti disampaikan oleh jaksa di Belanda.
\\
\midrule
\sc Full FT &  Seorang perempuan Belanda yang diduga telah bergabung dengan kelompok yang menamakan diri Negara Islam atau ISIS, telah menyelundup dua anak-anak ke Suriah, kata otoritas Belanda.
\\
\midrule
\sc LoRA-4 &Seorang perempuan Belanda yang diduga telah bergabung dengan kelompok jihadis ISIS di Suriah telah ditangkap di perbatasan Suriah.
\\
\bottomrule
\end{tabular}
}
\end{table*}

\clearpage

\begin{table*}[!ht]
\centering
\scalebox{0.8}{
\addtolength{\tabcolsep}{0pt}
\begin{tabular} {p{2cm}|p{16cm}}
\toprule
\textsc{Language} & \textsc{Korean}   \\
\midrule
\sc Input &  이곳은 오늘날 훨씬 건조한 곳이 됐다 이 지역은 현재 염전으로 활용되고 있지만, 20만 년 전 인간의 조상이 번성했을 당시엔 거대한 호수가 있었다. 인류의 조상은 기후 변화가 닥치기 전까지 7만 년 동안 그곳에 정착했다고 연구진은 설명했다. 비옥한 녹지가 다른 지역에도 확대돼 인류가 아프리카 바깥으로도 진출하기까지 그들은 후대 인류를 위해 길을 닦아놓았다. 호주 가반 의학연구소의 유전학자인 바네사 헤이스 교수는 "해부학적으로 약 20만 년 전 현대인류가 아프리카에 나타난 것은 어느 정도 분명한 사실"이라고 말했다. "오랫동안 논의해온 건 인류가 정확히 어디에서 기원했으며 어디서부터 퍼졌느냐는 것이다." 그러나 헤이스 교수의 결론에 같은 분야의 다른 학자들은 회의적인 반응을 보였다. 호수가 있는 안식처 문제의 지역은 남부 잠베지강 유역이다. 보츠나와 북부에 있는 곳이다. 보츠나와는 아프리카 대륙 남쪽에 자리했다. 연구원들은 인류의 조상이 거대한 호수 지역인 마카디카디 근처에 정착했다고 생각한다. 이 호수는 현재 소금 평면이 펼쳐져 있는 지역이다. 하이스 교수는 "여긴 매우 넓은 지역이었다. 과거엔 습지였을 수도 있고 풀이 우거졌을지도 모른다"라고 말했다. 그는 "현생 인류와 야생 동물에게 살기에 최적의 공간을 제공했을 것"이라고 덧붙였다. 이 지역에 7만 년 동안 살아온 인류는 이 지역에 강우량이 증가하면서 이주를 시작했다. 13만 년 전, 11만 년 전에 세 차례의 이주 물결이 있었다. 인류는 비옥한 초록색 토지가 있는 곳으로 퍼져나갔다. 나미비아의 칼라하리 사막에서 줄호안시 사냥꾼들에게 불을 피우는 법을 배우는 헤이스 교수 첫 이주민들은 북동쪽으로 향했다. 두 번째 이주 인파는 남서쪽으로 향했다. 3분의 1 정도만 이곳에 남아 지금까지 번성해오고 있다. 이주 인파를 연구하기 위해 현재 아프리카에 사는 수백 명의 미토콘드리아 DNA(어머니가 자식에게 물려주는 DNA 파편)를 이용하여 인류 가계도를 역추적했다. 유전학을 지리학과 기후 시뮬레이션에 결합함으로써, 연구진은 아프리카 대륙이 20만 년 전을 추정할 수 있었다. 인류 이야기를 다시 쓰다 그러나 이 연구는 네이쳐에 기고되자마자 곧장 반발에 맞닥뜨렸다. 다른 과학자는 미트콘드리아 DNA만 가지고 인류 역사를 재구성할 수 없다고 주장했다. 다른 연구들도 아프리카 동부에서 인류가 기원했음을 암시하는 화석을 발견하면서 다른 해답을 제시했다. 런던의 자연사 박물관 소속 크리스 스팅어 교수는 호모 사피엔스(현생 인류)의 진화는 복잡한 과정이라고 지적했다. 그는 BBC 뉴스에 "오늘날 인간의 미토콘드리아 분포만 활용해선 인류의 기원을 단일 지역으로 특정할 수 없다"고 말했다. "유전자의 아주 작은 부분만으로는 인류 기원의 모든 이야기를 들려줄 수 없기 때문에 (이 연구가) 데이터를 과도하게 활용하는 것이라고 생각한다." 결국 인류에게는 (아직 구체적으로 추정할 수 없는) 하나의 고향보다는 여러 곳의 고향이 있을 가능성도 있다. 인류 역사에서의 대표적인 진화 지표.
\\
\midrule
\sc Target &  과학자들은 현생 인류의 고향이 아프리카 잠베지 강 남부 지역이라고 추정한다.
\\
\midrule
\sc Full FT &  현재 아프리카에 사는 인류가 약 20만 년 전부터 살았던 곳이라는 연구가 나왔다.
\\
\midrule
\sc LoRA-4 & 아프리카에서 인류가 기원했다는 연구가 나왔다.
\\
\midrule
\midrule
\sc Input & 지난해, 코트렐은 모스크 건립에 항의한다는 뜻으로 모의 참수형을 하는 영상을 올려 유죄판결을 받았다 지난 5일 블레어 코트렐은 호주 스카이 뉴스의 일대일 스튜디오 인터뷰에 출연해 이민 관련 이야기를 했다. 그러나 시청자들은 코트렐이 지난해 이슬람교도들을 모욕한 죄로 유죄판결을 받았던 인물이라고 지적했다. 또, 코트렐이 아돌프 히틀러의 사진을 학교에 전시해야 한다고 주장했던 부분도 문제 삼았다. 결국 스카이 뉴스 호주는 인터뷰 당일, 트위터를 통해 "블레어 코트렐을 인터뷰한 것이 잘못된 것이었다"며 "그의 의견이 우리의 의견을 반영하는 것은 아니다"고 말했다. 해당 인터뷰를 온라인에서도 삭제했다고도 전했다. 코트렐은 이민 반대 단체인 애국전선연합(United Patriots Front)의 전 리더였다. 노던 테리토리 주의 수석 장관을 지낸 프로그램 진행자 애덤 자일스가 그를 인터뷰 했다. 같은 방송국 다른 앵커들도 이 인터뷰를 비난하고 나섰다. 앵커 로라 제이예스는 "블레어 코트렐은 자신이 히틀러 팬이라고 스스로 고백한 극우 파시스트다. 그는 여성을 조종하기 위해 '폭력과 테러'를 사용하는 것을 자랑했다"고 트위터에 글을 올렸다. 정규해설위원이자 호주 정부 장관을 역임했던 그래이그 에머스은 "우리나라에서 인종주의와 편협성을 일상화하는 여정에 발걸음을 내딘 것"이라고 했다. 그는 앞으로 이 방송국에 출연하지 않겠다고 선언했다. 코트렐은 호주 이민을 줄이고, 외국의 이념으로부터 보호하며, "우리의 전통적인 정체성"을 되찾기 위해 인터뷰를 이용했다고 했다. 그는 논란이 일자 스카이 뉴스가 '압력'에 굴복했다고 말했다. 지난해, 코트렐은 모스크 건립에 항의한다는 뜻으로 모의 참수형을 하는 영상을 올렸다. 그 결과 이슬람 교도들을 향해 혐오와 경멸을 조장했다는 혐의로 다른 2인과 함께 유죄 판결을 받았다. 지난해 다른 방송사인 채널 세븐 역시 이런 코트렐의 배경을 알리지 않고 생방송 인터뷰를 해 비난이 일기도 했다.
\\
\midrule
\sc Target &  호주의 한 TV 방송국이 극우 인사이자 과거 범죄 기록이 있는 인물과의 인터뷰한 후 호주 전역에서 분노가 일자, '잘못된 일'이었다고 말했다.
\\
\midrule
\sc Full FT &  호주 방송사가 히틀러 팬이라는 이유로 호주 이민자와의 인터뷰를 거부한 뒤 사과했다.
\\
\midrule
\sc LoRA-4 & 호주의 한 방송국이 히틀러를 존경하는 극우파 여성을 인터뷰한 것을 두고 비난이 쏟아지고 있다.
\\
\bottomrule
\end{tabular}
}
\end{table*}

\clearpage

\begin{table*}[!ht]
\centering
\scalebox{0.8}{
\addtolength{\tabcolsep}{0pt}
\begin{tabular} {p{2cm}|p{16cm}}
\toprule
\textsc{Language} & \textsc{Russian}   \\
\midrule
\sc Input &  Помпео сказал, что у него нет данных, указывающих на непосредственную угрозу ядерной войны с Северной Кореей Помпео, однако, признал, что оружейная программа Пхеньяна развивается темпами, которые вызывают "все большую тревогу", и заметил, что новые испытания северокорейских ракет не должны никого удивлять. В интервью Fox News он подчеркнул, что "стратегическое терпение", с которым Вашингтон взирал на ракетные и ядерные усилия Пхеньяна, закончилось. На вопрос о том, насколько близка Северная Корея к созданию ядерного оружия, способного достичь территории США, директор Центрального разведывательного управления ответил: "Теперь они ближе". "Я слышал тех, кто рассуждает о том, что мы стоим на пороге ядерной войны. Я сам не видел разведывательных данных, указывающих, что мы сегодня находимся в таком положении", - отметил Помпео. Президент США Дональд Трампа после публикаций в СМИ о том, что КНДР, вероятно, создала ядерный заряд, достаточно компактный, чтобы разместить его на имеющихся у нее баллистических ракетах, выступил с серией резких заявлений и твитов, в частности о том, что он готов ответить Пхеньяну "огнем и яростью" и что военные возможности для этого уже приведены в полную готовность. Северная Корея, со своей стороны, опубликовала план ракетного удара по острову Гуам в Тихом океане, где размещаются ключевые американские военные базы. Гуам: почему острову угрожает Северная Корея?
\\
\midrule
\sc Target & Непосредственной угрозы ядерной войны с Северной Кореей нет, несмотря на рост напряженности, заявил директор ЦРУ Майк Помпео.
\\
\midrule
\sc Full FT &  Директор ЦРУ США Майкл Помпео заявил, что Северная Корея "ближе" к созданию ядерного оружия, чем когда-либо.
\\
\midrule
\sc LoRA-4 & Директор ЦРУ Джеймс Помпео заявил, что у Вашингтона нет данных, указывающих на непосредственную угрозу ядерной войны с Северной Кореей.\\
\midrule
\midrule
\sc Input & Об этом заявила глава миссии Фиона Фрейзер во время представления 13-го отчета работы мониторинговой миссии. По словам представительницы ООН, еще 21 044 человека были ранены. Как заявила Фрейзер, это количество включает мирных граждан, военнослужащих и участников "различных вооруженных групп" на Донбассе. Мониторинговая Миссия ООН по правам человека начала свою работу в Украине в марте 2014 года. Миссия имеет офисы в Киеве, Донецке, Днепропетровске, Харькове, Краматорске и Одессе, которые покрывают всю зону, пострадавшую от конфликта, по обе стороны линии разграничения. Кроме того, Фиона Фрейзер также отметила, что мониторинговая миссия ООН обеспокоена возможным лишением социальных выплат вынужденных переселенцев в ходе проведения верификации таких выплат. "Мы обеспокоены возможным лишением социальных выплат переселенцев в рамках начатой недавно правительством проверки", - сказала она. Фрейзер заявила, что властям необходимо четко определить механизм такой проверки социальных выплат, чтобы не допустить нарушения прав переселенцев. Также она добавила, что правительству необходимо разработать эффективную программу компенсации для тех, кто потерял свое имущество в результате конфликта на Донбассе. Ранее министр социальной политики Павел Розенко заявил, что правительство прогнозирует в 2016 году экономию средств в сумме около пяти млрд гривен путем прекращения социальных выплат ненастоящим переселенцам, которые фактически проживают на неподконтрольных Киеву территориях Донецкой и Луганской областей.
\\
\midrule
\sc Target &  Мониторинговая миссия ООН по соблюдению прав человека в Украине объявила, что во время конфликта на Донбассе всего погибли 9167 человек.
\\ 
\midrule
\sc Full FT &  Мониторинговая миссия ООН по правам человека зафиксировала 10 000 погибших мирных жителей на Донбассе.
\\
\midrule
\sc LoRA-4 & В результате конфликта на востоке Украины за 12 месяцев 2015 года погибли 1 000 человек, 1 500 человек получили ранения, сообщило представительство ООН в Украине.
\\
\bottomrule
\end{tabular}
}
\end{table*}
\clearpage

\begin{table*}[!ht]
\centering
\scalebox{0.8}{
\addtolength{\tabcolsep}{0pt}
\begin{tabular} {p{2cm}|p{16cm}}
\toprule
\textsc{Language} & \textsc{Scottish Gaelic}   \\
\midrule
\sc Input &   Tha an t-ùghdarras ionadail air a bhith a' craoladh chèilidhean tron duilleag Facebook aca a h-uile oidhche Haoine on thòisich an glasadh-sluaigh sa Mhàrt. Bidh tè na bliadhn' ùire a' dol a-mach beò aig 19:00 air an aon làraich ma cheadaicheas na bacaidhean ionadail a bhios ann, agus bidh i cuideachd air a craoladh anns gach dachaigh-chùraim sna h-Eileanan Siar. Thuirt Co-òrdanaiche nam Meadhanan aig a' Chomhairle, Seòras Moireasdan: "Tha sinn air leth toilichte fàilte a chur air Willie Caimbeul, Iain 'Costello' MacÌomhair agus Calum Màrtainn, fir an taighe air oidhche na cèilidh. "Bidh iad a' seinn leotha fhèin agus còmhla. Bidh cuideachd òran ùr air a dhèanamh a dh'aona-ghnothaich airson na cuirme, le measgachadh de Bheurla agus Gàidhlig. "Gabhaidh an t-òran luchdachadh a-nuas às dèidh na cèilidh, agus thèid an t-airgead a thogar gu Hospice Bhethesda." Dùrachdan Thuirt Willie Caimbeul: "'S e bliadhna dhoirbh a th' air a bhith ann agus, mar sin, bha e mìorbhaileach a bhith a' seinn do dhaoine air na craolaidhean beò air-loidhne thairis air na sia mìosan a dh'fhalbh. "Tha mi air mo dhòigh gu robh mi an sàs ann agus tha mi a' coimhead air adhart gu mòr ri Cèilidh na Bliadhn' Ùire, àm nuair a tha cruinneachadh mar seo nas cudromaiche buileach. Tha spòrs gu bhith againn." Thuirt Calum Màrtain gur e urram a bh' ann cuireadh fhaighinn bho Chomhairle nan Eilean Siar nochdadh ann an dachaighean dhaoine còmhla ris na caraidean ciùil aige. "Tha e cho math gum bi cothrom agam a bhith air àrd-ùrlar còmhla ri dithis den luchd-ciùil as fheàrr air an eilean. Tha sinn an dòchas gun cuir sinn fiamh-ghàire air aodann dhaoine le òrain thraidiseanta ann an dà chuid Beurla agus Gàidhlig, agus tha sinn a' coimhead air adhart le dòchas ri 2021," thuirt Mgr Màrtainn. Cha robh Iain 'Costello' MacÌomhair misneachail an toiseach gun obraicheadh cèilidhean Covid na Comhairle beò, ach thuirt e gun do chòrd e ris gu mòr a bhith an sàs sa phròiseact a th' air togail agus taic a thoirt do dhaoine tron àm dhuilich. "Tha fiughair agam ri bhith ag obair còmhla ri dithis den luchd-ciùil agus seinneadairean as tàlantaiche as aithne dhomh agus tha mi air leth toilichte an cothrom seo a bhith agam a bhith a' seinn còmla ri Calum agus Willie." Bidh cothrom cuideachd aig a' phoball dùrachd a thoirt do chàirdean agus caraidean agus chithear na teachdaireachdan aca aig bonn an sgrion fhad 's a tha a' chèilidh a' dol air adhart. Bu choir do dhaoine an dùrachd a chur chun na comhairle air p.scott@cne-siar.gov.uk ro 17:00f Diciadain 25mh den t-Samhain. Thèid an còrr fiosrachaidh mun chèilidh fhoillseachadh sna beagan sheachdainnean ri thighinn.
\\
\midrule
\sc Target & Cumaidh Comhairle nan Eilean Siar Cèilidh Là na Bliadhn' Ùire air-loidhne agus cluinnear òran ùr aig a' chuirm airson a' chiad uair.
\\
\midrule
\sc Full FT & Gheibh Cèilidh na Bliadhna' Ùire 2021 air cuid de chuid de luchd-ciùil Chomhairle nan Eilean Siar ri fhailte beò air an 18mh Sultaine Diardaoin 18mh den Dùbhlachd.
\\
\midrule
\sc LoRA-4 & Tha Cèilidh na Bliadhn' Ùire na Chomhairle nan Eilean Siar a' dol a-mach beò an oidhche seo agus bidh tè na bliadhn' ùire a' dol a-mach beò air a' mhullach ann an Ceann a' Bhruthaidh.\\
\midrule
\midrule
\sc Input &Thuirt iad gum bu chòir do Riaghaltas na h-Alba a dhol an sàs ma thèid cead a thoirt seachad. Tha na h-Aimeireaganaich Todd Warnock is Mike Keiser airson raon aig àrd-ìre a thogail aig a' Chùil, eadar Eurabol agus Loch Fleòid. Chuir comhairlichean co-dhùnadh air a' chùis dheth nas tràithe air a' mhìos, ged a bha oifigich a' moladh a dhiùltadh. Tha an gnothach air ais aig a' Chomhairle Diciadain is oifigich a-rithist ag ràdh gun toireadh an leasachadh fìor dhroch bhuaidh air àrainneachd na sgìre. Tha dìon shònraichte air an sgìre timcheall Loch Fleòid. Rabhadh Tha Urras Fiadh-bheatha na h-Alba ag ràdh a-nise gur e seo "an cothrom mu dheireadh a' Chùil a ghlèidheadh". "'S e fìor ghlè bheag de dh'àiteachan san t-saoghal a tha air an dìon gu h-oifigeil mar a tha a' Chùil", thuirt Àrd-Stiùiriche an Urrais, Jonny Hughes. "Tha e gu sònraichte cudromach a thaobh nan dùintean-gainmhich mhachrach a th' ann, àrainneachd a tha ann an cunnart gu h-eadar-nàiseanta. "'Seo fear de na co-dhùnaidhean dealbhachaidh as cudromaiche ann an Alba anns na bliadhnaichean mu dheireadh. Mar sin, tha iomagain oirnn mu cho beag tuigse 's a sheall comhairlichean mu àrainneachd an àite is na beathaichean a tha a' fuireach ann", thuirt e. Dh'innis Mgr Hughes gu bheil e an dòchas gun tèid comhairlichean le moladh nan oifigeach. Thuirt e ge-tà, ma thèid cead a thoirt seachad, gum bu chòir do Riaghaltas na h-Alba an gnothach a ghairm a-steach cho luath 's a ghabhas. Bidh coinneamh shònraichte de Chomataidh Dealbhachaidh a Chinn a Tuath ann Diciadain gus meòrachadh air a' chùis.
\\
\midrule
\sc Target &  Dh'iarr Urras Fiadh-bheatha na h-Alba air comhairlichean na Gàidhealtachd iarrtas-dealbhachaidh 'son raon-goilf an Cataibh a dhiùltadh.
\\
\midrule
\sc Full FT &  Chuir oifigich aig Comhairle na h-Alba càineadh air sàrbhhearan 's iomagain air àrainneachd na h-àite is beathaichean a' Chùil Shiorrachd.
\\
\midrule
\sc LoRA-4 & Tha Urras Fiadh-bheatha na h-Alba ag ràdh gur e "an cothrom mu dheireadh a' Chùil a ghlèidheadh" gum faigh iad co-dhùnadh co-mhaoinnicheachail a thogail ann an sgìre Earra-Ghàidheal.
\\
\bottomrule
\end{tabular}
}
\end{table*}
\clearpage

\begin{table*}[!ht]
\centering
\scalebox{0.8}{
\addtolength{\tabcolsep}{0pt}
\begin{tabular} {p{2cm}|p{16cm}}
\toprule
\textsc{Language} & \textsc{Somali}   \\
\midrule
\sc Input &Madaxweyne Lula oo dhismaha uu ku jiro ka salaamaayo taageerayaashiisa. 72 sano jirkan ayaa lagu xukumay 12 sano oo xabsi ah, maadaama lagu helay eedo musuqmaasuq. Qareenadisa ayaa la sheegay in Booliska ay kala hadlayaan sidii uu Lula isu soo dhiibi lahaa. Waxaa soo baxaya warar sheegaya in Lula uu doonayo inuu booliska isu soo dhiibo maalinta sabtida. Siyaasigan ayaa horay ugu gacan seeyray amar maxkamadeed oo dhigayay in maalinta Jimcaha u booliska isku soo dhiibo. Taageerayaasha Lula oo isugu soo baxay banaanka dhismaha uu ku jiro. Kumanaan taageerayaashiisa ah ayaa isugu soo baxay banaanka dhismaha uu Lula Da Silva ku jiro. Mid kamid ah dadka isu soo baxay ayaa sheegay in haddi boolisku ay isku dayaan inay xiraan Lula aysan u suurtagelyn. Saraakisha ayaa hoosta ka xariiqay in madaxweynihii hore aan loo arkin qof baxsad ah, sida dadka qaar ba ay moodeen. Dibadbaxyo ka socda dalka Brazil Madaxweynaha Brazil oo dhaliilay waaxda caddaaladda Madaxwaynihii hore ee Brazil oo la amray in xabsiga la dhigo Muxuu Lula sidan u sameynayaa? Madaxweynihii hore ee dalka Brazil, Lula waxaa uu aaminsanyahay in xukunka ka dhanka ah isaga uu yahay mid siyaasadeysan. Lula ayaa sidoo kale ku doodaya in tallabaadani ay tahay qorshe lagu doonayo in lagaga hor istaago qorshaha uu ku doonayo in markale uu u tartamo xilka madaxtinimada ee dalkaasi oo uu sheegay inuu ku guuleysanayo. Qareenada Lula ayaa ku guul dareystay dadaalo ay ku doonayeen in Lula ay kaga badbaadiyaan xabsiga, daqiiqada ka hor xilligii kama dambeysta ah ee loo qabtay inuu isku soo dhiibo booliska.
\\
\midrule
\sc Target &   Madaxweynihii hore ee dalka Brazil Lula da Silva ayaa ku dhuumaaleysanaya dhismo ku yaalo magaalada uu ka soo jeedo ee ka baxsan Sao Paulo, kadib markii lagu amray inuu isku soo dhiibo ciidammada booliska.
\\
\midrule
\sc Full FT &  Madaxweynihii hore ee dalka Brazil, Luiz Inácio Lula da Silva, ayaa ka qeybgalaya xaaladda ka taagan xabsiga uu ku leeyahay magaalada Curitiba ee waqooyiga Brazil.
\\
\midrule
\sc LoRA-4 & Booliska Brazil ayaa madaxweynihii hore ee dalkaasi, Luiz Inacio Lula da Silva, ugu soo dhiibay xabsiga ee lagu xukumay inuu qabto xilka madaxtinimada.
\\
\midrule
\midrule
\sc Input & Gabadha ayaa aad u jeclayd bisadda Ruun waxa ay ku dadaashay sidii ay u heli lahayd cid u kabta bisadda, balse kuma aanay guulaysan oo waa ay ka bakhtiday. Gabadhaasi ayaa murugo badan ka qaaday geerida bisadda. Waxa ay xiriir la samaysan wasaaradda xanaannada xoolaha ee magaaladeeda balse kama aanay helin wax caawin ah. Sheekada Ruun iyo bisadeeda ayaa hadal hayn ka dhalisay baraha ay ku wada xiriiraan bulshada Soomaaliyeed, waxaana dadka qaar ay ku tilmaameen "qof waalan oo magac raadis ah." "Bisadeydu xitaa hilibka cayriinka ah ma aanay cuni jirin, oo waxaan usoo iibin jiray buskud iyo caano," ayay tiri Ruun. Balse gabadhan oo baratay culuunta caafimaadka dadweynaha ayaa ku doodaysa in haddii ay Soomaalidu u naxariistaan xayawaanka ay markaa heli karaan nabad. Hargeysa: Dhakhtar Yurub uga soo duulay si uu Harimacad u kabo Nin lagu xukumay inuu xabsiga ku dhex daawado Filim Mukulaal loo xukumay \$710 kun oo magdhow ah Ruun ayaa BBC-da laanta af Soomaaliga u sheegtay in hadda ka hor ay ka xanuunsatay bisadda, ka dibna ay la xiriirtay dhakhtar xayawaanka qaabilsan oo ku nool Masar, kaasoo kula taliyay dawada ku haboon bisadda. Waxa ay intaa ku dartay in ay rajeynayso in mustaqbalka ay furto goob ay ku xananaaneyso bisadaha, ayna diyaar u tahay inay raadsato bisad kale maadaama ay ka dhimatay tii ay jeclayd.
\\
\midrule
\sc Target &  Gabar ku nool magaalada Qardho oo lagu magacaabo Ruun Cali Cabdi ayaa waxa ay lahayd bisad ay aad u jeceshahay. Bisadii ayaa ka jabtay ka dib markii ay dhagax kaga dhufteen carruur yaryar.
\\
\midrule
\sc Full FT &   Ruun Cabdishakuur ayaa 26 jir ahayd markii ay 26-ka sano jirtay iyadoo xanuunsanaa bisadda lagu dilay xayawaanka caafimaadka dadweynaha ah ee lagu magacaabo 'Doleococcus'.
\\
\midrule
\sc LoRA-4 & Xayawaanka Ruun ayaa ku dhintay sidii ay u heli lahayd cid u kabta bisadda ee uu ku qabtay magaalada Hargeysa.
\\
\bottomrule
\end{tabular}
}
\end{table*}
\clearpage

\begin{table*}[!ht]
\centering
\scalebox{0.8}{
\addtolength{\tabcolsep}{0pt}
\begin{tabular} {p{2cm}|p{16cm}}
\toprule
\textsc{Language} & \textsc{Swahili}   \\
\midrule
\sc Input & Joao Teixeira de Faria, kati kati amejisalimisha kwa polisi Joao Teixeira de Faria, ambaye pia anafahamika kama 'Yohana wa Mungu' alitangazwa kuwa mtoro baada ya tarehe aliyopewa ya kujisalimisha kwa mamlaka kukamilika. Madai dhidi yake yalianza mapema Disemba wakati wanawake kadhaa walidai kuwa tabibu hiyo wa kiroho alikuwa amewanyanyasa kwenye kliniki yake. Bw Faira anakana madai hayo. Tabibu huyo ana makao yake mji ulio kati kati mwa nchi wa Abadiania magharibi mwa mji mkuu Brasilia, lakini ana wafuasi kote duniani. Alijisalimisha kwa njia gani? Gazeti la O Globo linasema kuwa dhehebu hilo lilitoa dola milioni 8.9 kutoka benki kadha siku ya Jumatano na kuashiria kuwa huenda alikuwa na njama ya kuikimbia Brazil au kuficha pesa hizo ikiwa labda atahitajika kulipa fidia. Tabibu wa kiroho Joao Teixeira de Faria anakana kuwanyanyasa wanawake Mamlaka zilijibu kwa kutangaza waranti wa kukamatwa siku ya Ijumaa. Siku ya Jumapili, video ya simu iliyopeperushwa na kituo cha televisheni cha Globo ilimuonyesha Bw Faria akitoka kwenye gari na kujisalimisha kwa polisi huko Abadiania. Alisafirishwa kwenda makao makuu ya polisi huko Goiania, mji mkuu wa jimbo Goias. Wakili wake Alberto Toron, alisema atakata rufaa Jumatatu. Alisema alikuwa na matumaini kuwa Bw Fariqa anaweza kuwekwa kwenye kifungo cha nyumbani badala ya jela. Madai dhidi yake yalianza vipi? Wiki iliyopita mpiga picha raia wa Uholanzi Zahira Leeneke Maus, aliiambia televisheni ya Globo kuwa Bw Faria alimshauri kufanya vitendo vya ngono na kisha kumbaka. Wanawake tisa raia wa Brazil ambao hawakutajwa majina, pia waliiambia televisheni hiyo kuwa tabibu huyo aliwanyanyasa kwa misingi kuwa alikuwa akiwapa nguvu za kuwasafisha. Televisheni ilionyesha Joao Teixeira de Faria akiwasili kituo cha polisi Baadaye gazeti la O Globo lilisema kuwa lilikuwa limezungumza na wanawake wawili zaidi ambao walikuwa na madai kama hayo. Wajumbe wa tabibu huyo mwenye miaka 76 wanasema anakataa madai ya kuhusika kwenye mambo yaliyo kinyume na sheria wakati akitoa matibabu. Bw Faria, ambaye mwenye si daktari, awali alipigwa faini na hata kufungwa kwa kuhudumu bila leseni. Mtangazaji wa runinga nchini Marekani Oprah Winfrey alisafiri kwenda nchini Brazil mwaka 2013 na kushuhudia akitoa matibabu kwenye kliniki yake.
\\
\midrule
\sc Target &  Tabibu wa kiroho nchini Brazil ambaye analaumiwa kwa kuwanyanyasa kingono zaidi ya wanawake 300 amejisalimisha kwa polisi, kwa mujibu wa vyombo vya habari.
\\
\midrule
\sc Full FT &  Tabibu mmoja wa kiroho nchini Brazil amejisalimisha kwa polisi baada ya madai ya kuwa aliwanyanyasa wanawake kwa kuwa aliwapa nguvu za kiume.
\\
\midrule
\sc LoRA-4 & Tabibu wa kiroho nchini Brazil amejisalimisha kwa polisi baada ya madai ya unyanyasaji na ukatili dhidi ya wanawake.
\\
\midrule
\midrule
\sc Input &Margaret Gieszinger alikamatwa baada ya video ya kisa hicho kusambazwa mtandaoni Margaret Gieszinger, 52, amefukuzwa kazi kwa kitendo hicho alichokifanya katika Shule ya Sekondari ya Chuo Kikuu cha huko Visalia, California. Waendesha mashtaka wamesema kuwa amekanusha makosa sita dhidi yake, likiwemo kosa la ukatili dhidi ya mtoto na shambulio la kimwili. Anakabiliwa na hukumu ya miaka mitatu unusu jela iwapo atakutwa na hatia. Bi Gieszinger, aliachiliwa kwa dhamana ya dola 100,000 Ijumaa jioni. Katika video hiyo ya iliyochukuliwa kwa simu ya mkononi kisha kupakiwa kwenye mtandao wa Reddit, mwalimu huyo anayefunza somo la sayansi, anaonekana akimuita mwanafunzi wa kiume kuketi mbele ya darasa, kisha akaanza kumkatakata nywele huku akiimba kwa makosa wimbo wa taifa la Marekani maarufu kama Star Spangled Banner. Wakili wa mwanafunzi huyo, ameliambia shirika la habari la CNN kuwa, mteja wake "alishtuka sana" kabla ya kufanikiwa kujinasua mikononi mwa mwalimu huyo. Bi Gieszinger kisha anaonekana kwenye video, akishika mkasi mkononi juu ya kichwa chake na kusema "next!" yaani "mwingine!" na kutishia pia kumkata nywele mwanafunzi mmoja wa kike. "Tunachukulia kwa tahadhari kubwa mno usalama wa wanafunzi madarasani," hiyo ni kwa mjibu wa taarifa kutoka kwa afisi kuu ya Elimu ya kaunti ya Tulare. "Tunachunguza taarifa zote tunazozipokea na tutachukua hatua kali mno na zinazohitajika dhidi ya wafanyakazi wetu walio na utovu wa nidhamu." Taarifa hiyo imeongeza.
\\
\midrule
\sc Target &  Mwalimu mmoja nchini Marekani anakabiliwa na mashtaka ya uhalifu, baada ya picha ya video kusambaa mitandaoni akimkata kwa lazima nywele mwanafunzi mmoja darasani, huku akiimba wimbo wa taifa wa Marekani.
\\
\midrule
\sc Full FT &  Mwalimu mmoja raia wa Marekani amekamatwa baada ya video kusambazwa mtandaoni, akimnyanyasa mwanafunzi wa kiume kwa kumkata nywele alipokuwa akipitia mtihani.
\\
\midrule
\sc LoRA-4 &Mwalimu wa shule ya sekondari nchini Marekani amehukumiwa miaka mitatu unusu jela baada ya video ya kisa hicho kusambazwa mtandaoni.
\\
\bottomrule
\end{tabular}
}
\caption{Examples of input, target, and PaLM 2-XXS generated summaries with full fine-tuning and LoRA-4 in XLSum, trained on all languages available (high-data regime).
}
\label{tab:examples}
\end{table*}

\end{document}